\documentclass{article}

\usepackage[letterpaper, margin=1.2in]{geometry}
\usepackage{parskip}

\usepackage{authblk}

\usepackage{graphicx}
\usepackage[utf8]{inputenc}
\usepackage[T1]{fontenc}
\usepackage{hyperref}
\usepackage{url}
\usepackage{booktabs}
\usepackage{amsfonts}
\usepackage{amsmath}
\usepackage{nicefrac}
\usepackage{microtype}
\usepackage{natbib}
\usepackage{placeins}
\usepackage{todonotes}
\usepackage{dsfont}
\usepackage[ruled,vlined]{algorithm2e}
\usepackage[caption=false]{subfig}
\usepackage{tikz}
\usetikzlibrary{positioning, fit, arrows.meta, backgrounds, shapes.geometric, shadows}
\usepackage[font=small,labelfont=bf]{caption}

\newcommand{\bvy}{\mathbf{y}}

\newcommand{\bvX}{\mathbf{X}}

\newcommand{\bvh}{\mathbf{h}}

\newcommand{\bvS}{\mathbf{S}}

\newcommand{\bvk}{\mathbf{k}}

\title{Joint Neural Posterior Estimation and Model Selection for Chemical Reaction Networks}
\author[1]{J. Arturo Esquivel}
\author[2,3,4]{Jorge A. Campos-Gonzalez-Angulo}
\author[5,6]{Sergio Pablo-García}
\author[5,7,8,9,10,11]{Alán Aspuru-Guzik}
\author[1,12]{Vianey Leos-Barajas}

\affil[1]{Department of Statistical Sciences, University of Toronto, Toronto, Ontario, Canada}
\affil[2]{Schools of Mathematics \& Physics and Chemistry \& Molecular Biosciences, The University of Queensland, St Lucia, Queensland, Australia}
\affil[3]{Australian Research Council Centre of Excellence in Quantum Biotechnology, St Lucia, Queensland, Australia}
\affil[4]{Queensland Quantum Decarbonisation Alliance, St Lucia, Queensland, Australia}
\affil[5]{The Matter Lab, University of Toronto, Toronto, Ontario, Canada}
\affil[6]{Institute of Micro and Nanotechnology}
\affil[7]{Departments of Chemistry, Computer Science, Materials Science \& Engineering, Chemical Engineering \& Applied Chemistry, and Institute of Medical Science, University of Toronto, Toronto, Ontario, Canada}
\affil[8]{Acceleration Consortium, Toronto, Ontario, Canada}
\affil[9]{Vector Institute for Artificial Intelligence, Toronto, Ontario, Canada}
\affil[10]{Canadian Institute for Advanced Research (CIFAR), Toronto, Ontario, Canada}
\affil[11]{NVIDIA, Toronto, Ontario, Canada}
\affil[12]{School of the Environment, University of Toronto, Toronto, Ontario, Canada}
\date{}

\begin{document}

\maketitle

\begin{abstract}
Simulation-based Bayesian inference for Chemical Reaction Networks (CRNs) must simultaneously solve two computationally demanding problems: discriminating between overlapping reaction pathways and estimating reaction rate constants. Classical approaches, such as Sequential Monte Carlo ABC, rely on handcrafted summary statistics and scale poorly across large parameter spaces. To overcome this, we introduce a joint amortized inference architecture. Our pipeline utilizes shared 1D-CNNs and a permutation-invariant Set Attention mechanism to learn low-dimensional summary statistics from variable-length, multi-experiment time-series data. These statistics simultaneously condition a model selection network to compute posterior model probabilities and model-specific conditional Invertible Neural Networks (cINNs) that estimate parameter posterior distributions. We demonstrate the practical utility of our approach on a real protodeboronation example, illustrating how it handles structural ambiguity and extracts kinetic parameters from competing catalyst deactivation pathways.
\end{abstract}

\section{Introduction}
\label{sec:intro}

Chemical reaction networks (CRNs) are at the heart of many societal and scientific concerns, such as designing efficient industrial chemical processes \citep{blau_chemically_2021}, optimizing pharmaceutical production \citep{dong_data-driven_2023}, and addressing climate-related challenges \citep{heald_fuel_2020}. Chemists decompose complex chemical processes into diagrams with elementary (or intermediate) reactions that occur over time. Their dynamics are dictated by reaction rate constants---quantitative measures of the propensity that an elementary reaction occurs within the network \citep{gillespie_stochastic_2007}. In practice, these underlying dynamics are studied by analyzing time-varying concentration data of the chemical species within the system.

A complete CRN describes all intermediate reactions that occurred within a specified time frame. Yet, chemical concentrations are often measured sparsely across time, and it is frequently only possible to measure a few of the chemical species involved in the process \citep{jovanovski_simulation-based_2025}. Consequently, different combinations of elementary reactions (CRN structures) can adequately describe similar concentration profiles (time-series). The scientific challenge lies in: (1) determining the valid graphical representation of the CRN that includes all elementary reactions; (2) robustly estimating the elementary reaction rates; and (3) inferring the changes that unmeasured species experience over time \citep{raue_structural_2009}.

Estimating the reaction rates is challenged by intractable chemical kinetics likelihood functions. Approximate Bayesian computation (ABC), and its extensions, is the standard for this kind of likelihood-free scenarios. But, it scales poorly when evaluating large parameter spaces and relies heavily on the choice of summary statistics and rejection thresholds. To conduct model selection, ABC typically relies on estimating Bayes Factors via expensive sequential Monte Carlo sampling \citep{toni_approximate_2008, didelot_likelihood-free_2011}, a process that scales poorly in the context of CRNs, as the networks complexities increase.

Recent breakthroughs, most notably by \cite{bures_organic_2023}, have demonstrated the utility of deep learning for studying reaction mechanisms (CRNs). Their approach trains a deep neural network to map kinetic data to mechanistic categories. While groundbreaking, they approach the problem solely as a classification task, and omit the continuous parameter space and associated uncertainty quantification. We generalize and expand upon this foundational idea. Our architecture utilizes normalizing flows, set attention, and a model neural network to: (1) quantify model uncertainty, learning calibrated posterior probabilities from chemical data; (2) estimate posterior distributions for the parameters governing the reactions involved in each model (CRN) considered; and (3) amortize the cost of inference, allowing for cheap evaluation of newly observed chemical data. We note that a similar architecture (combining Set Transformers with Convolutional Neural Networks and Normalizing flows) is concurrently being explored as a software tool for inference in electrochemistry \citep{yan_bing_spark_2026}. Our inference pipeline is tailored to accommodate time-series of variable length across multiple experimental conditions as commonly observed when studying kinetic chemical systems.

\section{Methods}\label{sec:method}

\subsection{Chemical Reaction Networks}\label{sec:crn}
Consider a network representing a chemical process composed of $n_r$ elementary reactions involving $n_s$ distinct chemical species. The dynamic evolution of this system is parameterized by a vector of reaction rate constants $\bvk \in \mathbb{R}_{+}^{n_r}$. For a well-mixed system with constant temperature, the state of the network at time $t$ is given by the $n_s$-dimensional vector of the number of molecules per species, $\bvX(t)$. The state changes of the network are encoded in the $n_s \times n_r$ stoichiometric matrix $\bvS$, where the entry $S_{ij}$ represents the net change in the number of molecules of species $i$ upon the occurrence of reaction $j$.

The rate at which these reactions occur is determined by an $n_r$-dimensional propensity vector, $\mathbf{h}(\mathbf{X}(t), \mathbf{k})$ \citep{gillespie_stochastic_2007}. Let $\mathbf{S}^-$ be the $n_s \times n_r$ reactant stoichiometric matrix, where each entry $S_{ij}^-$ specifies the number of molecules of species $i$ consumed by reaction $j$. Under mass-action kinetics, the propensity $h_j$ of the $j$-th reaction is proportional to the product of its available reactants:
\begin{equation}
\label{eq:propensity}
h_j(\bvX(t), \bvk) = k_j \prod_{i=1}^{n_s} X_i(t)^{S_{ij}^-}, \quad \text{for } j = 1, \dots, n_r.
\end{equation}
When the volume and the number of molecules are sufficiently large, stochastic fluctuations become marginal \citep{fearnhead_inference_2014} and changes in the system can be well approximated by deterministic Ordinary Differential Equations (ODEs):
\begin{equation}
\label{eq:ode}
\frac{d\bvX(t)}{dt} = \bvS\bvh (\bvX(t), \bvk).
\end{equation} 

We employ this ODE formulation as our primary simulator. To perform Bayesian inference on $\bvk$, evaluating the likelihood of the observed data requires numerically integrating these non-linear ODEs. Because this integration is computationally expensive, exact inference via Markov chain Monte Carlo (MCMC) quickly becomes prohibitive, as it demands sequential simulations for every proposed parameter step. Furthermore, the observable, typically consisting of multiple independent experiments $\bvy_0 = \{f(\bvX^{(1)}), \dots, f(\bvX^{(N_{\text{exp}})}) \}$, contains measurements from only a subset of the chemical species in the system. Because intermediate species are often not observed, the kinetic parameters $\bvk$ can be unidentifiable, requiring exhaustive exploration of the parameter space to accurately estimate their posterior distributions. 

\subsection{Classical Likelihood-Free Baseline}\label{sec:abc_baseline}
To bypass the problems of exact inference, Simulation-Based Inference (SBI) relies on generating synthetic datasets from the simulator. The most widely used traditional approach is Sequential Monte Carlo Approximate Bayesian Computation (SMC-ABC) \citep{sisson_sequential_2007}. SMC-ABC utilizes sequential importance sampling to iteratively filter proposed parameters $\bvk$ through a sequence of decreasing discrepancy tolerances $\epsilon$, comparing summary statistics $T(\bvy_{\text{sim}})$ against $T(\bvy_0)$. Moreover, by estimating the marginal likelihood from the importance sampling weights, SMC-ABC can approximate the posterior probability of model $M$, $p(M \mid \bvy_0)$, enabling selection between competing CRN structures via Bayes factors \citep{toni_approximate_2008, didelot_likelihood-free_2011}. An overview of the mathematical formulation, along with the algorithm used for our SMC-ABC baseline is provided in Appendix \ref{app:smc_abc}.

While theoretically sound, SMC-ABC still suffers three operational inefficiencies: (1) the curse of dimensionality requires handcrafted, low-dimensional summary statistics $T(\cdot)$ that are difficult to design and will likely discard relevant temporal kinetics \citep{fearnhead_constructing_2012}; (2) achieving precise inference and model selection also depends heavily on the schedule of discrepancy tolerances used \citep{robert_lack_2011}; and (3) much of the computational cost is incurred repeatedly for every newly observed dataset.

\subsection{Amortized Inference and Model Selection via Deep Learning}\label{sec:amortized-bayesflow}
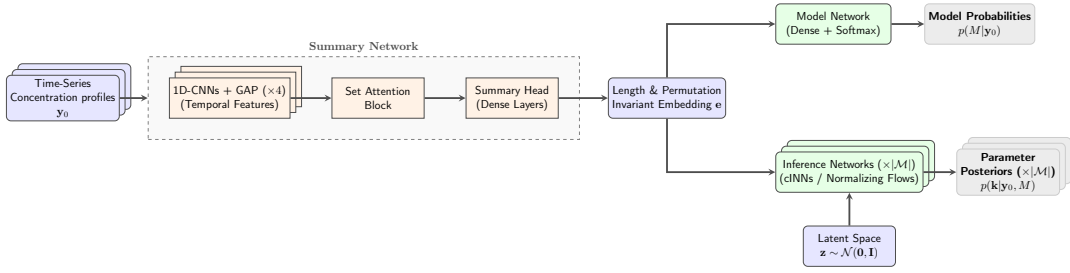
\begin{figure*}[htbp]
    \centering
      \resizebox{0.9\linewidth}{!}{%
      \begin{tikzpicture}[
          >=stealth,
          node distance=1.5cm and 1.5cm,
          every node/.style={align=center, font=\sffamily\normalsize},
          data/.style={rectangle, draw=black!70, thick, fill=blue!10, rounded corners, minimum height=3.5em, minimum width=8em},
          data_stack/.style={data, double copy shadow={shadow xshift=5pt, shadow yshift=5pt}},
          layer/.style={rectangle, draw=black!70, thick, fill=orange!10, minimum height=3.5em, minimum width=8em},
          layer_stack/.style={layer, double copy shadow={shadow xshift=5pt, shadow yshift=5pt}},
          network/.style={rectangle, draw=black!70, thick, fill=green!10, rounded corners, minimum height=3.5em, minimum width=10em},
          network_stack/.style={network, double copy shadow={shadow xshift=5pt, shadow yshift=5pt}},
          output/.style={rectangle, draw=gray!40, thick, fill=gray!15, rounded corners, minimum height=3.5em, minimum width=8em, font=\sffamily\normalsize\bfseries},
          output_stack/.style={output, double copy shadow={shadow xshift=5pt, shadow yshift=5pt}}
        ]
        
        \node[data_stack] (input) {Time-Series \\ Concentration profiles \\ $\bvy_0$};
        
        \node[layer_stack, right=1.5cm of input] (cnn) {1D-CNNs + GAP ($\times 4$) \\ (Temporal Features)};
        \node[layer, right=1.25cm of cnn] (attn) {Set Attention\\ Block};
        \node[layer, right=1.25cm of attn] (head) {Summary Head \\ (Dense Layers)};
        
        \begin{scope}[on background layer]
            \node[draw=black!50, dashed, thick, fill=gray!5, inner sep=18pt, fit=(cnn) (attn) (head), label={[font=\normalsize\bfseries, text=black!70]above:Summary Network}] (sumnet) {};
        \end{scope}
        
        \node[data, right=1.5cm of head] (embedding) {Length \& Permutation \\ Invariant Embedding $\mathbf{e}$};
        
        \node[network, above right=1cm and 1.5cm of embedding] (evidential) {Model Network \\ (Dense + Softmax)};
        \node[output, right=1cm of evidential] (pM) {Model Probabilities \\ $p(M | \bvy_0)$};
        
        \node[network_stack, below right=1cm and 1.5cm of embedding] (cinn) {Inference Networks ($\times |\mathcal{M}|$) \\ (cINNs / Normalizing Flows)};
        \node[output_stack, right=1cm of cinn] (pK) {Parameter \\ Posteriors ($\times |\mathcal{M}|$) \\ $p(\mathbf{k} | \bvy_0, M)$};
        
        \node[data, below=1cm of cinn] (latent) {Latent Space \\ $\mathbf{z} \sim \mathcal{N}(\mathbf{0}, \mathbf{I})$};
        
        \draw[->, ultra thick, draw=black!70] (input) -- (sumnet.west);
        \draw[->, ultra thick, draw=black!70] (cnn) -- (attn);
        \draw[->, ultra thick, draw=black!70] (attn) -- (head);
        \draw[->, ultra thick, draw=black!70] (head) -- (embedding);
        
        \draw[->, ultra thick, draw=black!70] (embedding) |- (evidential);
        \draw[->, ultra thick, draw=black!70] (evidential) -- (pM);
        
        \draw[->, ultra thick, draw=black!70] (embedding) |- (cinn);
        \draw[->, ultra thick, draw=black!70] (latent) -- (cinn);
        \draw[->, ultra thick, draw=black!70] (cinn) -- (pK);

      \end{tikzpicture}%
      }
      
      \caption{High-level architecture of the joint neural inference pipeline. Each time-series observation ($4$ in total) goes through a parallel, shared-weight CNN to extract kinetic features. The architecture bifurcates to evaluate model evidence and condition $\mathcal{M}$ independent model-specific conditional Invertible Neural Networks to estimate parameter posteriors.}
      \label{fig:macro_architecture}
\end{figure*}

Addressing the scaling limitations of traditional ABC, we designed a deep learning architecture leveraging the BayesFlow library \citep{kuhmichel_bayesflow_2026}. By framing Bayesian inference and model selection as a joint problem of representation learning, neural density estimation, and classification, we transfer the computational burden of simulating the CRN into an initial training phase. We achieve this through a unified architecture that tackles these tasks using three jointly trained components. First, a summary network \citep{radev_bayesflow_2022} extracts optimal statistics from the concentration time-series. Second, this learned representation conditions inference networks \citep{papamakarios_fast_2016} to estimate parameter posteriors. Third, the same representation simultaneously conditions a model network \citep{elsemuller_deep_2024} to compute posterior model probabilities. Together, these three components target the factorized joint posterior:
\begin{equation}
    p(M, \bvk \mid \bvy_0) = p(M \mid \bvy_0) \cdot p(\bvk \mid \bvy_0, M),
\end{equation} 
where $M \in \mathcal{M}$ represents a specific model (CRN pathway), $\bvk$ represents the kinetic parameters, and $\bvy_0$ denotes the observed concentration data. Figure \ref{fig:macro_architecture} shows a high-level overview of the architecture used.

\subsubsection{The Summary Network}
Circumventing the challenges involved in designing summary statistics, we define a summary network $T_\phi: \mathcal{Y} \to \mathbb{R}^{d_e}$ with learnable weights $\phi$. Because $T_\phi$ is trained jointly with the downstream classifier and density estimators, it is designed to preserve the mutual information between the concentration data and the target parameters, converging to a near-sufficient statistic \citep{radev_bayesflow_2022, chen_neural_2020}. 

Motivated by permutation-invariant set encoding \citep{zaheer_deep_2017}, the network treats the time-series from different experiments as an unordered set, ensuring the learned representation is independent of the order in which the results are provided. Each time-series is therefore passed independently through a shared one-dimensional Convolutional Neural Network (1D-CNN) to extract temporal dynamics \citep[e.g.,][]{ismail_fawaz_deep_2019, radev_bayesflow_2022}, followed by Global Average Pooling (GAP) that enforces length-invariance. This produces one fixed-length feature vector for each experimental setup. The collection of vectors is passed as an unordered set of tokens into a Set Transformer \citep{lee_set_2019} utilizing a Set Attention Block (SAB). Such mechanism allows the network to model interactions between different setups and assign contextual importance weights to the most informative conditions. The updated set of tokens is mean-pooled and passed through a Multi-Layer Perceptron (MLP) that yields a permutation-invariant summary embedding of fixed dimension $\mathbf{e} = T_\phi(\bvy_0)$. The detailed architecture is provided in Appendix \ref{app:network_architecture}.

\subsubsection{The Inference Network (Parameter Estimation)}
Because multi-modal posteriors and non-linear dependencies are prevalent in CRNs, we train an independent conditional Invertible Neural Network (cINN) for each candidate model $M \in \mathcal{M}$ \citep{radev_bayesflow_2022}. 

Each cINN $f_{\psi_M}(\bvk ; \mathbf{e})$ bijectively maps the reaction rate constants $\bvk$ to a latent variable $\mathbf{z} \sim \mathcal{N}(\mathbf{0}, \mathbf{I})$. Applying the change of variables theorem, the cINN defines a conditional probability density:
\begin{equation}
\label{eq:cinn_density}
p_{\psi_M}(\bvk \mid \bvy_0, M) = p_Z\left(f_{\psi_M}(\bvk ; \mathbf{e})\right) \left| \det \mathbf{J}_{f_{\psi_M}} \right|.
\end{equation}
Minimizing the expected forward Kullback-Leibler (KL) divergence between this approximate density and the true posterior $p(\bvk \mid \bvy_0, M)$ is equivalent to minimizing its expected negative log-density over the generative distribution \citep{radev_bayesflow_2022}:
\begin{align}
\label{eq:loss_param}
\mathcal{L}_{\mathcal{I}}(\phi, \psi_M) = \mathbb{E}_{p(\bvk, \bvy_0 \mid M)} \Big[ 
- \log p_Z\left(f_{\psi_M}(\bvk ; T_\phi(\bvy_0))\right) - \log \left| \det \mathbf{J}_{f_{\psi_M}} \right| \Big].
\end{align}

\subsubsection{The Model Network (Model Selection)}
Parameter inference assumes the underlying CRN structure is known. To jointly resolve this, the shared summary embedding $\mathbf{e}$ is also fed into a model network $g_\omega: \mathbb{R}^{d_e} \to \Delta^{|\mathcal{M}|}$. Implemented as an MLP with a softmax as the final layer, it maps the embedding $\mathbf{e}$ to a categorical distribution consisting of the posterior model probabilities $p_\omega(M \mid \bvy_0)$. 

The shared architecture approximates the joint posterior $p(M, \bvk \mid \bvy_0) = p(M \mid \bvy_0) p(\bvk \mid \bvy_0, M)$. Thus, minimizing the joint negative log-posterior reduces to minimizing the sum of the discrete and continuous losses:
\begin{equation}
\label{eq:nl-joint}
\begin{split}
- \log p(M, \bvk \mid \bvy_0) =  - \log p(M \mid \bvy_0) - \log p(\bvk \mid \bvy_0, M) .
\end{split}
\end{equation} 

Because the candidate models $M \in \mathcal{M}$ are discrete and mutually exclusive, the posterior $p(M \mid \bvy_0)$ corresponds to a categorical distribution. Consequently, the expected negative log-posterior in the first term expands into the standard categorical cross-entropy. Minimizing this objective---and assuming that the true data-generating process lies within the candidate set $\mathcal{M}$---makes the model network's softmax outputs converge to the true Bayesian posterior probabilities \citep{richard_neural_1991}. Substituting the model network's approximation $p_\omega$ and taking the expectation over the full generative process, the model loss becomes:
\begin{align}
\label{eq:loss_model}
\mathcal{L}_{\mathcal{M}}(\phi, \omega) = \mathbb{E}_{p(M, \bvk, \bvy_0)} \Big[ - \sum_{m \in \mathcal{M}} \mathbb{I}(M = m) \log p_\omega(M = m \mid T_\phi(\bvy_0)) \Big];
\end{align}
with $\mathbb{I}(\cdot)$ the indicator function. Hence, following the chain rule for KL divergence \citep{cover_elements_2006} and the factorization of the joint posterior, the global training objective we use to simultaneously address both tasks is the sum of the losses from Equations (\ref{eq:loss_param}) and (\ref{eq:loss_model}):
\begin{equation}
\label{eq:loss_total}
\mathcal{L}(\phi, \boldsymbol{\psi}, \omega) = \mathcal{L}_{\mathcal{I}}(\phi, \boldsymbol{\psi}) + \mathcal{L}_{\mathcal{M}}(\phi, \omega),
\end{equation}
where $\boldsymbol{\psi} = \{\psi_M\}_{M \in \mathcal{M}}$ is the collection of all model-specific cINN weights. We minimize the Monte Carlo estimate of (\ref{eq:loss_total}) across mini-batches of simulated triplets $(M, \bvk, \bvy_0)$, amortizing the computational cost of both tasks simultaneously. Once trained, this setup reduces joint inference to a forward pass, eliminating the need for further repetitive and expensive ODE simulations.

\section{Simulation Study}
\label{sec:simulation}

To validate and assess our proposed neural posterior estimation pipeline, we conducted a simulation study comparing it against the classical SMC-ABC baseline using synthetically generated CRN datasets with known ground-truth reaction mechanisms and parameter sets.

\subsection{Data and Problem Setting}
\label{sec:data}

We consider a model space $\mathcal{M}$ consisting of a finite set of candidate CRN topologies drawn from \citet{bures_organic_2023}. Our primary study focuses on three distinct reaction pathways, $\mathcal{M} = \{M_1, M_2, M_3\}$, assuming a uniform categorical prior $p(M)$. Each model $M \in \mathcal{M}$ is characterized by a parameter vector of reaction rate constants $\bvk_M \in \mathbb{R}_{+}^{d_M}$ governing its elementary reactions. To reflect an uninformative setting where no prior kinetic information is available, individual rate constants are drawn independently from log-uniform priors spanning multiple orders of magnitude, specifically $\log k_{i} \sim \mathcal{U}(\log 10^{-5}, \log 10^5)$ for all $k_i \in \bvk_M$.

Following the training framework of \citet{bures_organic_2023}, we consider $N_{\text{exp}} = 4$ independent experimental setups per system. To perturb the system and expose different kinetic regimes, each setup is initialized with varying starting catalyst concentrations, uniformly sampled between $1$ and $10$ \text{mol}\%. For each mechanism, our forward ODE simulator generated $936,000$ independent training trajectories and $1,000$ test trajectories. Generating these datasets required approximately $22$ wall-clock hours per mechanism utilizing $120$ CPU cores each. The ODE systems for the mechanisms and complete simulation details are provided in Appendix \ref{app:simulation_details}.

The observable $\bvy_0$ comprises time-series concentration measurements of a subset (Substrate and Product) of the involved chemical species. For a given experimental setup $i$, the continuous trajectory is sampled at $T=20$ time points. To mirror realistic laboratory conditions, these observations are distorted with measurement noise. The data at each time step $t$ is generated as $\bvy_0^{(i)}(t) \sim \mathcal{N}(\bvX^{(i)}_0(t), \Sigma)$, where $\bvX^{(i)}_0(t)$ represents the subset of measurable species derived from the simulator at time $t$, and $\Sigma = \text{diag}(\sigma_S^2, \sigma_P^2)$. The species-specific noise scale parameters, $\sigma_S$ and $\sigma_P$, are sampled from a log-normal prior and inferred jointly with the kinetic parameters (from $6$ to $8$ total parameters, depending on the model).

Including training (performed using a GPU), the three-model architecture took roughly $25$ hours to be ready for inference. Once trained, however, evaluating new experimental trajectories takes only a few seconds. Detailed hardware configurations are provided in Appendix \ref{app:compute}.

\subsection{Baseline Comparison: SMC-ABC}
\label{sec:baseline}

To establish a baseline, we initially implemented the SMC-ABC algorithm to evaluate discrepancies directly on point-wise observed concentrations via the $L_2$ distance across the observation vectors concatenated. We executed it using a population size of $3,000$ particles for each candidate model. We evaluated multiple threshold schedules, including adaptive thresholding \citep{lenormand_adaptive_2013}, where the discrepancy $\epsilon$ for a given generation is dynamically set to a specific quantile of the accepted discrepancies from the previous generation.

Evaluating discrepancies directly over the observation vectors proved inefficient. In classical ABC, working with high-dimensional data spaces is affected by the curse of dimensionality, as the probability of a simulated trajectory falling within an $\epsilon$-neighborhood of that observed diminishes exponentially \citep{sisson_handbook_2018}. In our setting, this is compounded by the uninformative priors and the varying parametric complexities of the candidate mechanisms, causing high rejection rates and thus making the implementation very costly. In an attempt to accelerate inference and improve posterior convergence, we augmented the concentration trajectories with additional summary features, such as local trajectory gradients. Yet, incorporating these statistics offered no meaningful improvement in either acceptance rates or model selection accuracy. Further tuning of these features could theoretically improve the performance. However, such problem-specific calibration requires significant manual effort and is exactly what deep learning approaches aim to circumvent.

To demonstrate how these limitations manifest in practice, we consider a representative test case (where the true generative mechanism was $M_2$). We implemented an adaptive threshold SMC-ABC algorithm across the candidate models, initialized with $\epsilon_0 = 8.5$ and an adaptive threshold quantile of $\alpha = 0.55$. For this one case, the algorithm required evaluating $928,605$ ODE simulations to produce $3,000$ samples for each model across $6$ generations. Furthermore, this baseline often failed to confidently, or even correctly, recover the true model. In this case, the highest estimated posterior probability belonged to $M_1$ ($53.1\%$), as opposed to the true mechanism $M_2$ ($46.8\%$). ABC inherently penalizes parametric complexity when drawing from broad priors, and so it tends to favor simpler mechanisms that can loosely mimic the data. Visual inspection of the posterior predictive trajectories (App. \ref{app:results_abc}, Figure \ref{fig:smc_abc_ppc}) shows that both models can produce data within the vicinity of the observed curves. Ultimately, this approach fails to distinguish between structurally distinct mechanisms that produce similar kinetic profiles. By contrast, our neural architecture is optimized to extract the higher-order temporal correlations and cross-species dependencies that can help make this distinction. 

This high computational cost makes classical simulation-based approaches, such as SMC-ABC, impractical for chemical systems, which often require repeated inference across varying experimental conditions. Executing SMC-ABC for just a single case like this requires almost one-third of the total simulations used to train our amortized framework. As demonstrated in the following section, by shifting this simulation cost to a one-time training phase, our neural architecture not only enables fast inference at test time, but also achieves high model selection performance.

\subsection{Amortized Neural Inference Evaluation}

Having established the computational limitations of classical SMC-ABC for this kind of chemical system, we evaluate the performance of our proposed neural posterior approach. We assessed the pipeline on an independent test set of 3,000 concentration trajectories (split equally across the three mechanisms), evaluating its performance across three dimensions: (1) model selection accuracy; (2) the calibration of the estimated posterior probabilities, and (3) the quality of parameter inference within the selected reaction pathway.

\subsubsection{Model Selection Accuracy}
\label{sec:model_selection}

To evaluate our approach's model selection capabilities, we measured how frequently it assigned the highest posterior probability to the true underlying mechanism. We assessed this accuracy under four distinct observational settings: (1) both the substrate and product are observed across all $T=20$ time points; (2) both species are observed at $T=10$ time points; (3) only the product is observed across all $T=20$ time points; and (4) only the product is observed at $T=10$ time points. Global accuracy metrics and per-model F1-scores for all four settings are summarized in Table \ref{tab:ablation_results}.

\begin{table*}[h]
\centering
\caption{Model Selection Performance of Our Deep Learning Pipeline. We evaluate it across varying observational conditions: the impact of hidden chemical species (Product Only) and reduced temporal resolution (10 vs. 20 points). Global accuracy and model-specific F1-scores are reported for the $3,000$ independent test trajectories.}
\label{tab:ablation_results}
\begin{tabular}{@{}llcccc@{}}
\toprule
\textbf{Observed Species} & \textbf{Time Points} & \textbf{Global Acc.} & \textbf{$M_1$ F1} & \textbf{$M_2$ F1} & \textbf{$M_3$ F1} \\
\midrule
Substrate \& Product      & 20                   & 94\%                 & 0.94              & 0.98              & 0.96              \\
Substrate \& Product      & 10                   & 94\%                 & 0.93              & 0.97              & 0.93              \\
Product Only              & 20                   & 93\%                 & 0.91              & 0.97              & 0.92              \\
Product Only              & 10                   & 92\%                 & 0.89              & 0.95              & 0.90              \\
\bottomrule
\end{tabular}
\end{table*}

Under the richest data scheme (substrate and product measured at $20$ time points), our network achieved a global model selection accuracy of $94\%$. This already contrasts with the classical SMC-ABC baseline, which struggled to confidently identify the true mechanism under various hyperparameter specifications. When the number of measurements was reduced to $10$ randomly sampled time points, the network demonstrated robustness, maintaining the $94\%$ global accuracy. 

Further, we tested the pipeline when assuming the substrate was unobserved and that only measurements of the product were available. We trained the very same architecture using only product trajectories, and used it to infer the mechanism strictly from the product's formation trajectory. Remarkably, the network retained its discriminatory power, achieving $93\%$ accuracy with $20$ observation points and $92\%$ with $10$. 

The confusion matrices across all four observation settings are shown in Appendix \ref{app:results_3model}, Table \ref{tab:classification_3model}. Mechanism $M_2$ proved to be the most distinguishable pathway, obtaining F1-scores between $0.96$ and $0.98$ across all observational settings. The network struggled more between $M_1$ and $M_3$, specifically when the substrate was hidden. For example, in the most constrained condition ($10$ product-only points), the network misclassified $M_1$ as $M_3$ in $71$ instances ($7.1\%$ of the cases), and $M_3$ as $M_1$ in $99$ instances ($9.9\%$ of the cases). This aligns with the underlying reaction kinetics: in $M_3$, by design the off-cycle resting state is only significant during initial conversion ($<10\%$), after which its product formation trajectory closely resembles that of $M_1$. Conversely, $M_2$ features a continuous separation of the active catalyst into an off-cycle dimer during the main half of the reaction, causing a distinct deceleration on the product curve that the network can identify even when the substrate is unobserved. Ultimately, these results confirm that the neural architecture successfully extracts the dynamics required to discriminate between reaction pathways, without critical reductions in performance when the underlying physical observables are decreased.

\subsubsection{Posterior Calibration}
\label{sec:calibration}

In real-world chemical systems, multiple reaction pathways often occur simultaneously. While one primary mechanism usually dominates the process, alternative pathways frequently operate in the background, meaning experimental data is rarely the product of a single, perfectly isolated rule \citep[see Chapter 2 of][]{steinfeld_chemical_1998}. Hence, and even though the network is trained assuming that there is only one true underlying model, a robust inference framework must do more than simply predict the most likely mechanism; it must produce well-calibrated posterior probabilities that correctly quantify model selection uncertainty. 

\begin{figure}[h!]
    \centering
    \includegraphics[width=.6\linewidth, trim={0.1cm 0.1cm 0.1cm 0.1cm}, clip]{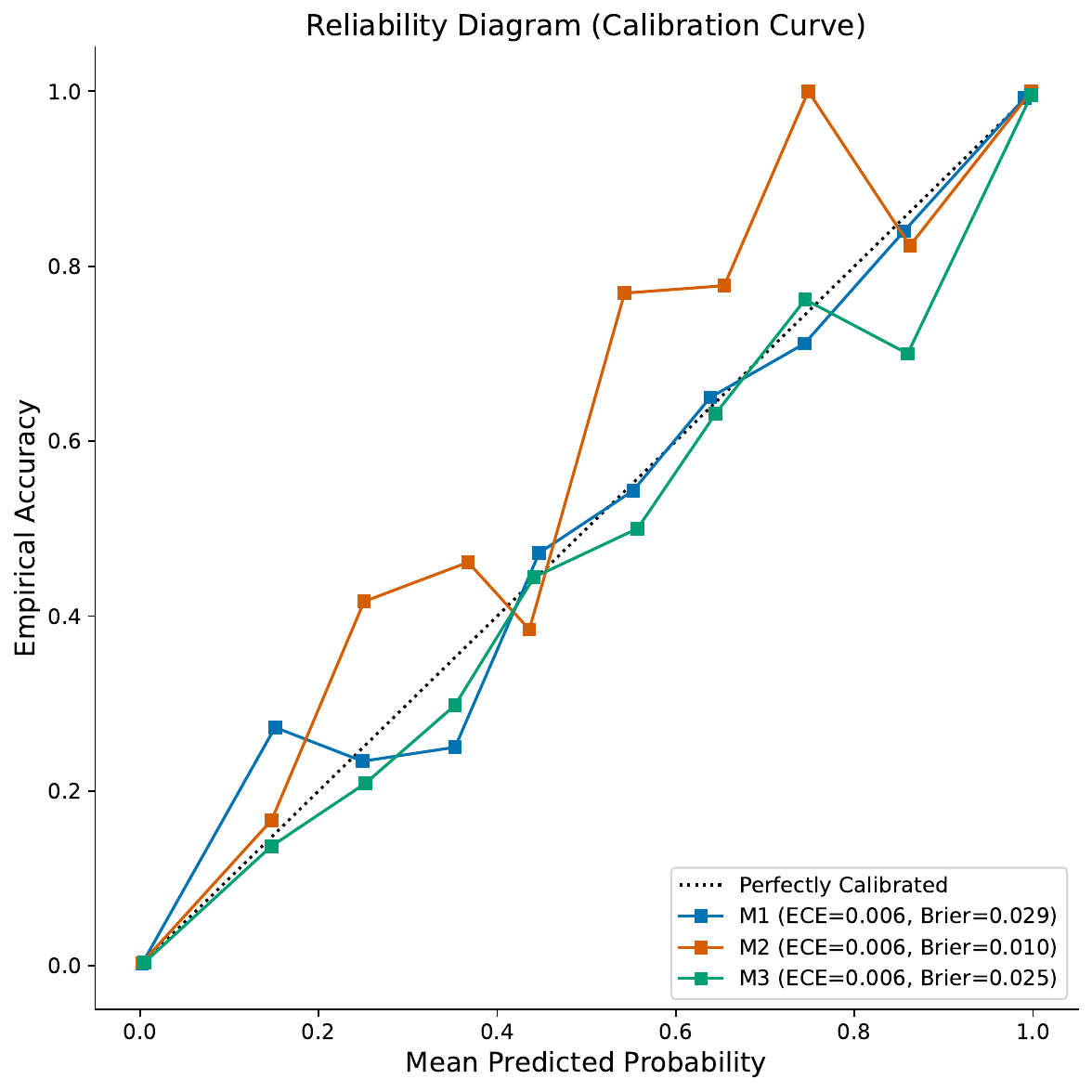}
    \caption{Reliability diagram evaluating the calibration of the posterior probabilities across the 3,000 test trajectories. Predictions are grouped into 10 uniformly spaced probability bins. The dashed diagonal represents perfect calibration, where the estimated probability exactly matches empirical accuracy. All three mechanisms closely track the identity line, achieving low Expected Calibration Errors.}
    \label{fig:calibration}
\end{figure}

To assess the uncertainty estimates from our network, we computed the Expected Calibration Error (ECE) using 10 uniform bins \citep{guo_calibration_2017} and Brier scores \citep{brier_verification_1950}. The network achieves low ECE values for all models: around $0.006$. This means the network's estimated posterior probabilities deviate from the testing accuracy by less than $1\%$ on average. Further, we produced reliability diagrams for all three mechanisms across the test set (under observational scheme (1) above). As shown in Figure \ref{fig:calibration}, the estimated posterior probabilities closely track the identity line. This indicates that the network's uncertainty closely matches its accuracy---when the network expresses a specific level of certainty (posterior probability), the frequency with which the true mechanism is correctly identified is very close to its estimated probability. While there are minor deviations in the intermediate probability bins, with slight underconfidence about $M_2$ and some overconfidence about $M_3$, these deviations occur only for a handful of bins and are generally not too large (less than $0.20$ in all cases but one). In general, our results suggest that our pipeline produces well-calibrated posterior probabilities. 

\subsubsection{Parameter Inference}
\label{sec:parameter_inference}

Having demonstrated the pipeline's ability to classify generative mechanisms, we next evaluate its accuracy in inferring reaction rate constants and generating parameter posteriors that reliably capture estimation uncertainty.

To measure this, we computed the empirical coverage for $95\%$ credible intervals (CIs) from the posterior of each kinetic parameter within the test set. In a perfectly calibrated Bayesian setup, $95\%$ of the ground-truth parameters should fall within the network's estimated CIs. Table \ref{tab:coverage_main} details the obtained coverage for the parameters of $M_1$, $M_2$, and $M_3$ under the richest observational setting (setup (1)). The network achieves close to perfect coverage across all parameters.

\begin{table}[h]
    \centering
    \caption{Empirical Coverage of the $95\%$ Posterior Credible Intervals. Results for the kinetic parameters and observation noise ($\sigma$) under the $20$-point (Substrate + Product) setting.}
    \label{tab:coverage_main}
    \begin{tabular}{lccc}
        \toprule
        \textbf{Parameter} & \textbf{$M_1$} & \textbf{$M_2$} & \textbf{$M_3$} \\
        \midrule
        $k_1$    & $96.2\%$ & $96.2\%$ & $96.2\%$ \\
        $k_{-1}$ & $96.4\%$ & $95.2\%$ & $96.0\%$ \\
        $k_2$    & $98.1\%$ & $97.3\%$ & $95.1\%$ \\
        $k_{-2}$ & $97.0\%$ & $94.8\%$ & $96.8\%$ \\
        $k_3$    & --       & $93.0\%$ & $95.2\%$ \\
        $k_{-3}$ & --       & $96.2\%$ & $96.0\%$ \\
        $\sigma_S$ & $95.1\%$ & $93.0\%$ & $95.8\%$ \\
        $\sigma_P$ & $95.1\%$ & $93.0\%$ & $94.3\%$ \\
        \bottomrule
    \end{tabular}
\end{table}

Moreover, the architecture maintains high-quality parameter uncertainty even when observables are restricted. Under the most constrained setting (product at $10$ time points), the $95\%$ CI coverage remains high across all parameters, ranging between $89.5\%$ and $97.7\%$. Full coverage results for all observational settings are provided in Appendix \ref{app:results_3model}, Table \ref{tab:master_coverage_3model}. 

Additionally, we performed Simulation-Based Calibration (SBC) \citep{talts_validating_2020} to verify the probabilistic calibration of the marginal posteriors. The resulting SBC rank histograms (shown in App. \ref{app:results_3model}, Figures \ref{fig:rank_M1_1}, \ref{fig:rank_M2_1} and \ref{fig:rank_M3_1}) exhibit uniformity across the majority of the parameters. Further, it shows how the network handles weakly identifiable parameters. For example, the rate $\log k_{-1}$ in $M_3$ cannot be fully recovered from the observables, with the posterior mean estimates often being very similar for different true values. Its posterior is more informative than the prior (the variance shrinkage was $\approx 0.26$), but still spans most of the prior space. However, its corresponding SBC rank histogram remains close to uniform. This indicates that the network does not produce incorrect, overconfident posteriors, and instead outputs a suitably uncertain distribution. Minor deviations do exist in some cases---as for $\log k_3$ in $M_2$. But the overall uniformity indicates that the cINNs produce well-calibrated marginal posteriors on top of the underlying model selection.

Finally, we conducted posterior predictive checks (PPCs) to ensure these posteriors translate into sensible concentration profiles. To capture both model and parameter uncertainty, we first sample a mechanism using the estimated posterior probabilities, and then draw rate constants from its corresponding posteriors, using them to produce 100 simulations. As shown in the example in Figure \ref{fig:ppc_1} (App. \ref{app:results_3model}), the estimated parameters yield predictive distributions that closely track the ground-truth trajectories. This indicates that the network successfully captures the true underlying chemical kinetics, while accounting for parameter and model uncertainties given the data observed.

\subsubsection{Scalability: Four Candidate Models}
\label{sec:scalability}
Unlike classical methods (e.g., SMC-ABC) where inference costs increase significantly with the number of candidate models, our amortized approach can scale more efficiently. We expanded the model space to include a fourth mechanism, $M_5$ (App. \ref{app:M5}), simply by adding a classifier output and a corresponding cINN. 

Despite keeping the network's parameter capacity fixed (e.g., the summary network dimensions), global classification accuracy across $4,000$ test trajectories remains high at $86\%$ (ECE $\leq 0.0229$). We expect that scaling the hidden layers would easily offset this minor performance drop.

Parameter recovery also handles the expanded space robustly. Although $M_5$ introduces weakly identifiable parameters (e.g., $\log k_{-2}$ exhibits a low variance shrinkage of $0.080$), the $95\%$ credible intervals maintain acceptable coverage ($95.3\%$--$97.8\%$) across all its eight parameters (App. \ref{app:results_4model}). This indicates that the network accurately quantifies uncertainty when faced with uninformative physical signals. 

Moreover, while adding a new mechanism requires one-time upfront simulation and training ($\approx 25.8$ hours; Section \ref{sec:simulation}), the inference time for new experimental trajectories remains amortized and is instantaneous. Thus, the architecture supports rapid, repetitive inference over complex model spaces without the computational inefficiencies of classical approaches.

\section{Applied Example: Protodeboronation}
\label{sec:real_world}

As a proof of concept on a real-world example, we applied our framework to experimental kinetic data from a catalytic protodeboronation---an undesirable degradation process in Suzuki-Miyaura cross-coupling \citep{ser_bulky_2025}. To identify the dominant mechanism driving this degradation, we considered three competing mechanisms for catalyst deactivation from \citet{bures_organic_2023}: a unimolecular deactivation of the catalyst-substrate complex ($M_{14}$), an inhibitor-induced deactivation ($M_{15}$), and a substrate-induced deactivation where a second substrate neutralizes the active intermediate ($M_{16}$) (see App.\ \ref{app:real_world_params}).

We first trained our pipeline on synthetic simulations generated from these three candidate mechanisms. We then processed the experimental data from \citet{ser_bulky_2025}, which consisted of product concentration profiles sampled at $8$ time points across $N_{\text{exp}} = 3$ independent setups with varying initial catalyst loadings ($5\%$, $10\%$, and $15\%$). The network assigned a $92.19\%$ probability to $M_{15}$, considering it a stronger candidate than the alternative unimolecular ($M_{14}: 3.71\%$) and substrate-induced ($M_{16}: 4.10\%$) degradation hypotheses. This result provides actionable insights; further experimentation could focus on purifying the reaction environment to remove the deactivating agent.

\begin{figure*}[htbp]
    \centering
    \includegraphics[width=\linewidth, trim={0.25cm 0.2cm 0.7cm 0.2cm}, clip]{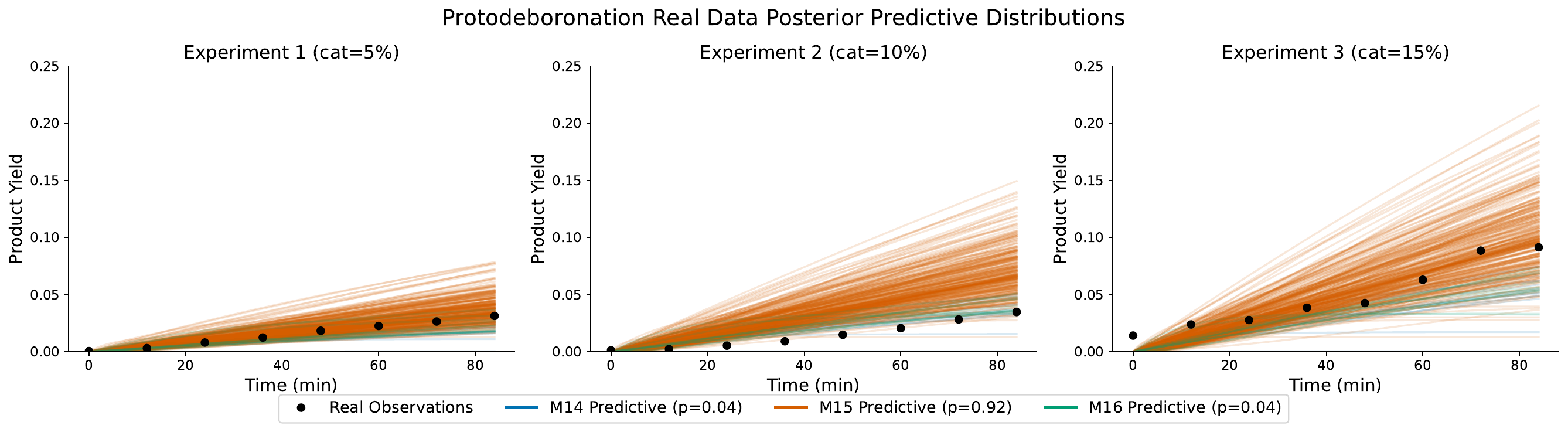}
    \caption{Posterior Predictive Check (PPC) for experimental protodeboronation data across three varying initial catalyst loadings. The $300$ predictive trajectories are sampled from the joint posterior, drawing from the candidate mechanisms proportionally to their estimated model probabilities.}
    \label{fig:real_data_ppc}
\end{figure*}

The inference network produced wide posteriors for most parameters (shown alongside point estimates in App. \ref{sec:real_results}), indicating that measuring the product alone is insufficient to fully identify all individual reaction rate constants. However, as shown in Figure \ref{fig:real_data_ppc}, synthetic trajectories drawn from these posteriors are consistent with the experimental observations. The network accurately maps covariant relationships between the parameters compatible with the true dynamics---as shown by the bivariate posterior distributions in App.\ \ref{app:real_world_params}, Figure \ref{fig:bivariate_pos}---while establishing their underlying uncertainty. By capturing these relationships, the pipeline reveals the underlying kinetic constraints of the system, offering deeper mechanistic insights even when absolute rate constants remain non-identifiable.

\section{Discussion}
\label{sec:discussion}

In this work, we introduced a joint neural posterior architecture for CRNs that combines reaction pathway selection and kinetic parameter estimation. We leverage a permutation-invariant summary network and couple it with cINNs and an MLP with a classification head to overcome the scaling and calibration limitations of classical likelihood-free methods like SMC-ABC. As demonstrated with synthetic cases and on real protodeboronation data, the pipeline extracts the necessary kinetic dynamics to output well-calibrated posterior model probabilities and robust parameter credible intervals, even when used on sparse observations.

The current framework was developed to consider three competing reaction mechanisms and was successfully scaled to a four-model space with minimal modifications and little performance loss. A natural next step is scaling the architecture to accommodate a larger number of competing structures. For example, \citet{bures_organic_2023} formalize 20 commonly encountered reaction mechanisms. Scaling the joint architecture to accommodate this magnitude of structures will likely require expanding the capacity of the summary network and potentially exploring hierarchical classification heads to deal with several overlapping structures, a direction we are actively pursuing.

Furthermore, our framework currently relies on a closed-world assumption: it presumes the true data-generating process lies within the predefined candidate set $\mathcal{M}$. If the true mechanism is absent from the candidate space, or if none of the formalized models accurately explain the experimental data, the model network is forced to allocate its probability mass across the available approximations. In practice, this limitation is partially mitigated by chemical intuition, as candidate structures are typically curated by domain experts before inference begins. Nevertheless, moving toward open-world inference remains a critical challenge. Equipping the architecture with the capability to flag an experimental trajectory as an anomaly within the candidates considered is a critical direction for future research.

By shifting the heavy computational burden of forward ODE simulations to a pre-inference training phase, our approach amortizes the cost of joint inference and enables a fast feedback loop between experimentation and robust analysis. This could pave the way for integration into automated laboratories, enabling real-time mechanism discovery and experimental design when studying complex chemical systems.

\section*{Code Availability}
The source code for the joint neural inference architecture, implemented via BayesFlow, along with the scripts required to simulate the chemical trajectories and reproduce all experiments, are publicly available at \url{https://github.com/Esquivel-Arturo/JNPE}.

\section*{Acknowledgements}

J.A.E. is supported by a Doctoral Student Fellowship from the Data Sciences Institute at the University of Toronto. J.A.E. and V.L.-B. acknowledge the support of the Canadian Statistical Sciences Institute (CANSSI) and CANSSI Ontario. V.L.-B. also acknowledges the support of the Natural Sciences and Engineering Research Council of Canada (NSERC).  J.A.C.G.A. acknowledges funding of this project by the NSERC Alliance Grant \#ALLRP587593-23 (Quantamole). J.A.C.G.A. also acknowledges support from the Council for Science, Technology and Innovation (CSTI), Cross-ministerial Strategic Innovation Promotion Program (SIP), “Promoting the application of advanced quantum technology platforms to social issues” (Funding agency: QST). J.A.C.G.A. also acknowledges support from the Australian Research Council Centre in Quantum Biotechnology through project number CE230100021. S.P.-G. acknowledges financial support from the Comunidad de Madrid through the “César Nombela” talent attraction program (Grant No. 2025-T1/COM36440)

\bibliographystyle{plainnat}
\bibliography{references} 

\clearpage
\appendix

\begin{center}
    \LARGE \textbf{Joint Neural Posterior Estimation and Model Selection for Chemical Reaction Networks: \\ Supplementary Materials}
\end{center}
\vspace{0.5cm}

\section{Sequential Monte Carlo ABC Formulation}
\label{app:smc_abc}

A comprehensive theoretical treatment of Approximate Bayesian Computation (ABC) is beyond the scope of this work (we refer the reader to standard literature such as \citet{toni_approximate_2008}, \citet{beaumont_approximate_2002} and \citet{sisson_sequential_2007}). Here we briefly outline the SMC-ABC algorithm utilized as our classical baseline to provide intuition for its computational mechanics.

Instead of evaluating the likelihood directly, ABC simulates synthetic data $\bvy_{\bvk}$ using candidate parameters and compares it to the observed data $\bvy_0$. A candidate parameter set is accepted if the distance between their respective summary statistics falls below a specified tolerance threshold: $\rho(T(\bvy_{\bvk}), T(\bvy_0)) \leq \epsilon$, where $T(\cdot)$ is a summary statistic function and $\rho(\cdot, \cdot)$ is a distance metric (e.g., Euclidean distance).

To avoid the inefficient rejection rates of naive ABC rejection sampling, SMC-ABC makes use of sequential importance sampling. A population of $N$ particles is propagated through a sequence of intermediate distributions defined by decreasing tolerances $\epsilon_1 > \dots > \epsilon_R$. At generation $r$, a candidate particle $\bvk^{**}$ is generated by sampling an ancestor $\bvk^*$ from generation $r-1$ with probability proportional to its importance weight $w^{(r-1)}$, and perturbing it via a Gaussian transition kernel $q(\cdot \mid \bvk^*)$.

If the simulated data satisfies the current tolerance $\rho(T(\bvy_{\bvk^{**}}), T(\bvy_0)) \leq \epsilon_r$, the particle is accepted. To account for the proposal distribution, the unnormalized importance weight for the accepted particle $i$ is defined as:
\begin{equation}
\label{eq:smc_weight}
w^{(r)}_i = \frac{p(\bvk_i^{(r)})}{\sum_{j=1}^N w^{(r-1)}_j q\left(\bvk_i^{(r)} \mid \bvk_j^{(r-1)}\right)}.
\end{equation}

Under this framework, estimating the marginal likelihood (evidence) for a specific mechanism $M$ relies on a Monte Carlo approximation over all $N_{\text{total}}$ proposed particles. ABC approximates the likelihood through $\mathbb{I}(\rho(T(\bvy_{\bvk}), T(\bvy_0)) \leq \epsilon_R)$, which evaluates to $1$ if the simulated data falls within the tolerance and $0$ otherwise. 

By substituting this indicator function into the standard importance sampling estimator, the sum over all proposals reduces to only the $N_{\text{acc}}$ accepted particles. Thus, the evidence estimate reduces to:
\begin{equation}
\label{eq:evidence_approx}
p(\bvy_0 \mid M) \approx \frac{1}{N_{\text{total}}} \sum_{i=1}^{N_{\text{total}}} \frac{\mathbb{I}(\rho_i \leq \epsilon_R) p(\bvk_i \mid M)}{\mu_R(\bvk_i \mid M)} \propto \frac{1}{N_{\text{total}}} \sum_{i=1}^{N_{\text{acc}}} w_R^{(i)},
\end{equation}
where $\mu_R(\bvk \mid M) = \sum_{j} \bar{w}_j^{(R-1)} q(\bvk \mid \bvk_j^{(R-1)})$ is the proposal density, and $w_R^{(i)}$ the unnormalized weights at the final generation $R$.

Therefore, for a finite set of competing mechanisms $\mathcal{M}$, selecting between hypotheses $M_i$ and $M_j$ via the posterior odds ratio involves:
\begin{equation}
\label{eq:bayes_factor}
\frac{p(M_i \mid \bvy_0)}{p(M_j \mid \bvy_0)} = \frac{p(M_i)}{p(M_j)} \underbrace{\frac{p(\bvy_0 \mid M_i)}{p(\bvy_0 \mid M_j)}}_{K_{ij}} \approx \frac{p(M_i)}{p(M_j)} \frac{\sum w_R^{(i, M_i)} / N_{\text{total}}^{(M_i)}}{\sum w_R^{(j, M_j)} / N_{\text{total}}^{(M_j)}}.
\end{equation}

Note that this corresponds exactly to the Bayes Factor $K_{ij}$ when the model prior is uniform.

Since the marginal likelihood estimate in Equation \ref{eq:evidence_approx} requires tracking the total number of simulations $N_{\text{total}}$, the SMC-ABC pipeline is executed independently for every candidate model $M \in \mathcal{M}$. As the tolerance $\epsilon_R \to 0$, $N_{\text{total}}$ grows exponentially, making classical Bayes Factor estimation computationally prohibitive for large hypothesis spaces. The procedure used to execute this independent model evaluation (inspired by implementations from \citet{warne_simulation_2019} and \citet{didelot_likelihood-free_2011}) is detailed in Algorithm \ref{alg:smc-abc-resampling}.

\begin{algorithm}[h!]
\SetAlgoLined
\KwData{Observed data $\bvy_0$, Prior $p(\bvk)$, Simulator $\mathcal{S}(\bvk, \mathcal{R})$, Metric $\rho(\cdot, \cdot)$, Particles $N$, Tolerances $\epsilon_1 > \dots > \epsilon_R$.}
\KwResult{Posterior samples $\{\bvk_i^{(R)}\}_{i=1}^N$ approximating $p_{\epsilon_R}(\bvk \mid \bvy_0)$.}

\vspace{0.2cm}
\tcp{Initialization (Generation $r=1$)}
\For{$i \gets 1$ \KwTo $N$}{
    \Repeat{$\rho(\bvy_{\bvk^{**}}, \bvy_0) \leq \epsilon_1$}{
        Sample $\bvk^{**} \sim p(\bvk)$\;
        Simulate $\bvy_{\text{sim}} \sim \mathcal{S}(\bvk^{**}, \mathcal{R})$\;
    }
    Set $\bvk_i^{(1)} \gets \bvk^{**}$\;
    Set $w_i^{(1)} \gets 1/N$\;
}

\vspace{0.2cm}
\tcp{Sequential Importance Resampling}
\For{$r \gets 2$ \KwTo $R$}{
    \For{$i \gets 1$ \KwTo $N$}{
        \Repeat{$\rho(\bvy_{\bvk^{**}}, \bvy_0) \leq \epsilon_r$}{
            Sample ancestor $j$ from $\{1, \dots, N\}$ with prob $w^{(r-1)}_j$\;
            Sample proposed parameter $\bvk^{**} \sim q(\bvk \mid \bvk_j^{(r-1)})$\;
            Simulate $\bvy_{\bvk^{**}} \sim \mathcal{S}(\bvk^{**}, \mathcal{R})$\;
        }
        Set $\bvk_i^{(r)*} \gets \bvk^{**}$\; 
        $w_i^{(r)*} \gets \frac{p(\bvk_i^{(r)*})}{\sum_{j=1}^N w_j^{(r-1)} q(\bvk_i^{(r)*} \mid \bvk_j^{(r-1)})}$\;
    }
    
    \vspace{0.1cm}
    \tcp{Normalize and Resample}
    $W \gets \sum_{i=1}^N w_i^{(r)*}$\;
    \For{$i \gets 1$ \KwTo $N$}{
        $w_i^{(r)*} \gets w_i^{(r)*} / W$\;
    }
    \For{$i \gets 1$ \KwTo $N$}{
        Sample index $m$ with prob $w^{(r)*}_m$\;
        Set $\bvk_i^{(r)} \gets \bvk_m^{(r)*}$\;
        Set $w_i^{(r)} \gets 1/N$\;
    }
}
\caption{SMC-ABC with Sequential Importance Resampling}
\label{alg:smc-abc-resampling}
\end{algorithm}

\newpage
\section{Neural Architecture and Training Details}

\subsection{Deep Learning Architecture}
\label{app:network_architecture}

This section details the neural network architectures and training hyperparameters used for the simulation studies and applied example. The framework was implemented using TensorFlow/Keras and the BayesFlow library.

The architecture consists of three core components: (1) a Permutation-Invariant CNN-Transformer summary network that compresses the multivariate concentration trajectories $\bvy$ into a fixed-length embedding; (2) a model selection network utilizing a dense body; and (3) a group of conditional Invertible Neural Networks (cINNs) for model-specific parameter inference. A detailed visual diagram of the summary network is shown in Figure \ref{fig:macro_architecture}. The configurations and training hyperparameters for the entire architecture are summarized in Table \ref{tab:hyperparameters}.

The summary network employs a permutation-invariant structure. First, the 1D-CNN and the initial dense projection layer are applied independently to each of the $N$ experimental observations (time-series) using shared weights. The data is then reshaped and passed into the Self-Attention block, which allows the observation embeddings to contextualize one another. Finally, a global average pooling layer across the observations combines the interacting sequences into the representation that is passed to the shared dense body. 

The output of this shared body is then fed into a dedicated classification head, which outputs logits via a linear activation. During inference, a Softmax function is applied to these logits to yield the final posterior model probabilities $p(M | \bvy_0)$. 

In parallel with the model selection task, the shared summary embedding $\mathbf{e}$ conditions parameter inference. We instantiate a dedicated cINN for each model. Each cINN is constructed as a normalizing flow comprising a sequence of 8 affine coupling layers \citep{ardizzone_conditional_2019}. Each coupling block is parameterized by internal MLPs configured with two hidden layers of 256 units (BayesFlow's default). Each independent cINN learns to map a standard Gaussian to the joint posterior distribution $p(\bvk \mid \bvy_0, M)$ for the specific kinetic parameters of that mechanism.

\begin{table}[ht]
\centering
\caption{Architecture and Training Hyperparameters for our Neural Inference Framework.}
\label{tab:hyperparameters}
\begin{tabular}{@{}ll@{}}
\toprule
\textbf{Component / Hyperparameter} & \textbf{Specification} \\ \midrule
\multicolumn{2}{@{}l}{\textit{Summary Network: CNN-Transformer}} \\ 
1D-CNN filters (3 layers, kernel size 3) & $64 \rightarrow 128 \rightarrow 256$ \\
Pooling Layer & Global Average Pooling (across time steps, masked for dropout points) \\
Dense layer ($d_{model}$) & $256$ \\
Self-Attention heads & $4$ \\
Feed Forward Network & $512 \rightarrow 256$ \\
Normalization & Add \& Norm (LayerNorm) \\
Pooling layer & Global Average Pooling (across experimental observations) \\
Dense layers & $512 \rightarrow 256$ \\
Final embedding & $256$ \\
Activation functions & ReLU \\
Dropout rate & $0.1$ \\ \midrule
\multicolumn{2}{@{}l}{\textit{Model Network: MLP}} \\ 
Dense layer (classification head) & $256 \rightarrow 3$ (base) / $256 \rightarrow 4$ (four models) \\
Activation function & Linear \\ \midrule
\multicolumn{2}{@{}l}{\textit{Inference Networks: cINNs}} \\ 
Coupling design & Affine \\
Number of coupling blocks & $8$ \\
MLP hidden layers per block & $2$ \\
MLP hidden units per layer & $256$ \\
Inference parameters ($M_1, M_2, M_3, M_5$) & $6, 8, 8, 8$ \\ \midrule
\multicolumn{2}{@{}l}{\textit{Training Setup}} \\ 
Optimizer & Adam \\
Learning rate & $ 1 \times 10^{-4}$ (Constant) \\
Gradient clipping (norm) & $1.0$ \\
Batch size (per mechanism) & $180$ \\
Total samples per mechanism & $936,000$ \\
Total training epochs & $55$ \\ \bottomrule
\end{tabular}
\end{table}

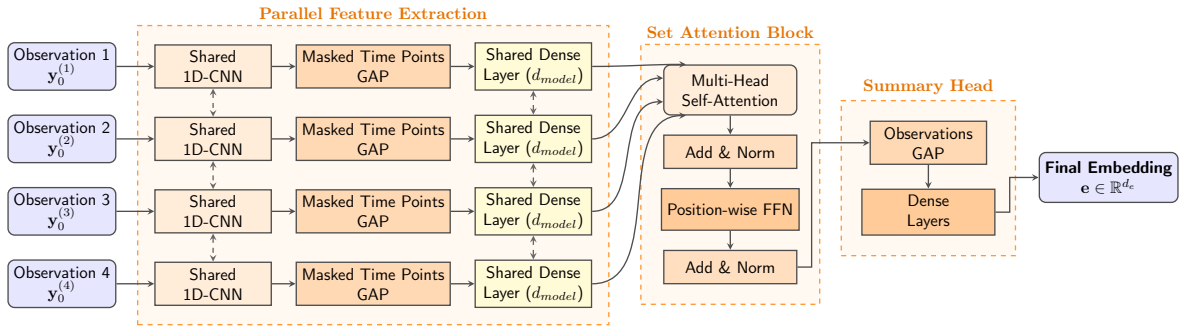
\begin{figure}[ht]
    \centering
    \resizebox{\textwidth}{!}{%
    \begin{tikzpicture}[
        >=stealth,
        node distance=1.2cm and 1.2cm,
        every node/.style={align=center, font=\sffamily},
        input/.style={rectangle, draw=black!70, thick, fill=blue!10, rounded corners, minimum height=2.5em, minimum width=6em},
        cnn/.style={rectangle, draw=black!70, thick, fill=orange!15, minimum height=2.5em, minimum width=7em},
        gap/.style={rectangle, draw=black!70, thick, fill=orange!30, minimum height=2.5em, minimum width=7em},
        proj/.style={rectangle, draw=black!70, thick, fill=yellow!20, minimum height=2.5em, minimum width=7em},
        attn/.style={rectangle, draw=black!70, thick, fill=orange!15, rounded corners, minimum height=3.0em, minimum width=8em},
        norm/.style={rectangle, draw=black!70, thick, fill=orange!25, minimum height=2.0em, minimum width=8em},
        ffn/.style={rectangle, draw=black!70, thick, fill=orange!40, minimum height=2.5em, minimum width=8em},
        output/.style={rectangle, draw=black!70, thick, fill=blue!10, rounded corners, minimum height=3em, minimum width=7em, font=\sffamily\bfseries}
      ]
      
      \node[input] (in1) {Observation 1 \\ $\mathbf{y}_0^{(1)}$};
      \node[input, below=0.5cm of in1] (in2) {Observation 2 \\ $\mathbf{y}_0^{(2)}$};
      \node[input, below=0.5cm of in2] (in3) {Observation 3 \\ $\mathbf{y}_0^{(3)}$};
      \node[input, below=0.5cm of in3] (in4) {Observation 4 \\ $\mathbf{y}_0^{(4)}$};
      
      \node[cnn, right=0.8cm of in1] (cnn1) {Shared \\ 1D-CNN};
      \node[cnn, right=0.8cm of in2] (cnn2) {Shared \\ 1D-CNN};
      \node[cnn, right=0.8cm of in3] (cnn3) {Shared \\ 1D-CNN};
      \node[cnn, right=0.8cm of in4] (cnn4) {Shared \\ 1D-CNN};
      
      \draw[<->, dashed, thick, draw=black!60] (cnn1.south) -- (cnn2.north);
      \draw[<->, dashed, thick, draw=black!60] (cnn2.south) -- (cnn3.north);
      \draw[<->, dashed, thick, draw=black!60] (cnn3.south) -- (cnn4.north);
      
      \node[gap, right=0.5cm of cnn1] (gap1) {Masked Time Points \\ GAP};
      \node[gap, right=0.5cm of cnn2] (gap2) {Masked Time Points \\ GAP};
      \node[gap, right=0.5cm of cnn3] (gap3) {Masked Time Points \\ GAP};
      \node[gap, right=0.5cm of cnn4] (gap4) {Masked Time Points \\ GAP};

      \node[proj, right=0.5cm of gap1] (proj1) {Shared Dense \\ Layer ($d_{model}$)};
      \node[proj, right=0.5cm of gap2] (proj2) {Shared Dense \\ Layer ($d_{model}$)};
      \node[proj, right=0.5cm of gap3] (proj3) {Shared Dense \\ Layer ($d_{model}$)};
      \node[proj, right=0.5cm of gap4] (proj4) {Shared Dense \\ Layer ($d_{model}$)};

      \draw[<->, dashed, thick, draw=black!60] (proj1.south) -- (proj2.north);
      \draw[<->, dashed, thick, draw=black!60] (proj2.south) -- (proj3.north);
      \draw[<->, dashed, thick, draw=black!60] (proj3.south) -- (proj4.north);
      
      \begin{scope}[on background layer]
          \node[draw=orange!80, dashed, thick, fill=orange!5, inner sep=10pt, fit=(cnn1) (cnn4) (gap1) (gap4) (proj1) (proj4), label={[font=\bfseries, text=orange!90!black]above:Parallel Feature Extraction}] (extraction) {};
      \end{scope}
      
      \path (proj2.east) -- (proj3.east) coordinate[midway] (proj_mid);
      
      \node[attn, right=1.5cm of proj_mid, yshift=1.8cm] (attention) {Multi-Head \\ Self-Attention};
      \node[norm, below=0.4cm of attention] (norm1) {Add \& Norm};
      \node[ffn, below=0.4cm of norm1] (ffn_block) {Position-wise FFN};
      \node[norm, below=0.4cm of ffn_block] (norm2) {Add \& Norm};
      
      \begin{scope}[on background layer]
          \node[draw=orange!80, dashed, thick, fill=orange!5, inner sep=12pt, fit=(attention) (norm1) (ffn_block) (norm2), label={[font=\bfseries, text=orange!90!black]above:Set Attention Block}] (transformer) {};
      \end{scope}

      \path (norm1.east) -- (ffn_block.east) coordinate[midway] (trans_mid);
      \node[gap, right=1.5cm of trans_mid, yshift=0.8cm] (pooling) {Observations \\ GAP};
      \node[ffn, below=0.5cm of pooling] (densebody) {Dense \\ Layers};
      
      \begin{scope}[on background layer]
          \node[draw=orange!80, dashed, thick, fill=orange!5, inner sep=12pt, fit=(pooling) (densebody), label={[font=\bfseries, text=orange!90!black]above:Summary Head}] (summaryhead) {};
      \end{scope}

      \path (pooling.east) -- (densebody.east) coordinate[midway] (head_mid);
      \node[output, right=1.0cm of head_mid] (embedding) {Final Embedding \\ $\mathbf{e} \in \mathbb{R}^{d_e}$};
      
      \draw[->, thick, draw=black!70] (in1) -- (cnn1);
      \draw[->, thick, draw=black!70] (in2) -- (cnn2);
      \draw[->, thick, draw=black!70] (in3) -- (cnn3);
      \draw[->, thick, draw=black!70] (in4) -- (cnn4);
      
      \draw[->, thick, draw=black!70] (cnn1) -- (gap1);
      \draw[->, thick, draw=black!70] (cnn2) -- (gap2);
      \draw[->, thick, draw=black!70] (cnn3) -- (gap3);
      \draw[->, thick, draw=black!70] (cnn4) -- (gap4);

      \draw[->, thick, draw=black!70] (gap1) -- (proj1);
      \draw[->, thick, draw=black!70] (gap2) -- (proj2);
      \draw[->, thick, draw=black!70] (gap3) -- (proj3);
      \draw[->, thick, draw=black!70] (gap4) -- (proj4);
      
      \draw[->, thick, draw=black!70] (proj1.east) to[out=0, in=180] (attention.150);
      \draw[->, thick, draw=black!70] (proj2.east) to[out=0, in=180] (attention.170);
      \draw[->, thick, draw=black!70] (proj3.east) to[out=0, in=180] (attention.190);
      \draw[->, thick, draw=black!70] (proj4.east) to[out=0, in=180] (attention.210);
      
      \draw[->, thick, draw=black!70] (attention) -- (norm1);
      \draw[->, thick, draw=black!70] (norm1) -- (ffn_block);
      \draw[->, thick, draw=black!70] (ffn_block) -- (norm2);
      
      \draw[->, thick, draw=black!70] (norm2.east) -- ++(0.3,0) |- (pooling.west);
      
      \draw[->, thick, draw=black!70] (pooling) -- (densebody);
      
      \draw[->, thick, draw=black!70] (densebody.east) -- ++(0.3,0) |- (embedding.west);
      
    \end{tikzpicture}%
    }
    \caption{Detailed architecture of the summary network $T_\phi$. Independent experimental observations are processed through a shared 1D-CNN (indicated by dashed connections) to extract temporal signatures. The sequences are globally average pooled across time to ensure length invariance. The independent features are contextualized using a Set Attention Block (SAB), before being globally pooled across experiments and passed through dense layers to produce the unified embedding $\mathbf{e}$.}
    \label{fig:summary_net}
\end{figure}

\subsection{Input Representation and Data Handling}

To process the multi-experiment trajectories, the input to the network is structured as a tensor of shape $(B, N_{\text{exp}}, T, C)$, where $B$ is the batch size, $N_{\text{exp}} = 4$ is the number of experimental initial conditions, and $T = 21$ represents the maximum number of temporal observations (including initial conditions). The feature dimension $C = 5$ consists of: the observation time $t$, the measured substrate concentration $[S]$, the measured product concentration $[P]$, the initial catalyst loading for that specific experiment $[cat]_0$, and a binary indicator flagging the initial condition $t=0$.

To ensure the summary network learns robust length-invariant representations that can generalize to varying experimental measurement rates (e.g., $T=20$ versus $T=10$), we apply temporal dropout during training. For each forward pass, we uniformly sample between $5$ and $T$ valid time points per experimental setup. The discarded time points are masked by setting their time channel to $-1.0$. All target parameters are standardized prior to density estimation.

\subsection{Data Flow}

The input tensor is processed through a sequence of transformations designed to extract temporal and contextual features:
\begin{itemize}
\item \textbf{Temporal Modeling (1D-CNN):} The $N_{\text{exp}}$ observation sequences are flattened into a batch of independent trajectories and passed through the shared 1D-CNN. The sequence length is convolved into $128$ and then $256$ feature channels.
\item \textbf{Masked Temporal Pooling:}  For each independent observation, the CNN features are averaged across the valid time steps—ignoring any points removed by temporal dropout—to collapse the sequence into a single vector. A shared dense layer then projects these independent vectors into a $(B, N_{\text{exp}}, d_{\text{model}})$ tensor.
\item \textbf{Set Interaction (SAB):} To capture cross-dependencies between the varying initial conditions, the tensor is processed by a Multi-Head Self-Attention block ($4$ heads, $d_{\text{model}}=256$), followed by Add \& Norm and Position-wise Feed-Forward layers.
\item \textbf{Summary Embedding:} The context-aware set of $N_{\text{exp}}$ vectors is mean-pooled across the experimental dimension. A final dense body maps this pooled vector to the fixed summary embedding $\mathbf{e} \in \mathbb{R}^{256}$.
\item \textbf{Inference and Model Networks:} The embedding $\mathbf{e}$ is passed through a linear classification head to output the categorical model logits, while simultaneously acting as the conditional input for the independent cINNs.
\end{itemize}

The architecture is trained end-to-end by minimizing a joint loss function. At each step, the total loss is computed as the sum of the categorical cross-entropy from the Model Network ($- \log p(M \mid \mathbf{y}_0)$) and the negative log-posterior from the corresponding cINN ($- \log p(\mathbf{k} \mid \mathbf{y}_0, M)$). The network weights are updated using the Adam optimizer with a constant learning rate of $1 \times 10^{-4}$ and gradient clipping scaled to a maximum norm of $1.0$ to prevent overflow during the early training stages. The model is trained over $55$ epochs using batch sizes of $180$ per mechanism. All other details and hyperparameter specifications are summarized in Table \ref{tab:hyperparameters}.

\subsection{Software Infrastructure and Computational Resources}
\label{app:compute}

To generate the training datasets efficiently, simulations were separated from the neural network training. The chemical systems were numerically integrated using Julia's \texttt{DifferentialEquations.jl} suite, accessed via the \texttt{diffeqpy} wrapper in Python. This enabled rapid, multi-core dataset generation performed on an academic high-performance computing cluster.

Our neural posterior inference pipeline was implemented in Python utilizing TensorFlow and the BayesFlow library \citep{kuhmichel_bayesflow_2026}. Generating the $936,000$ synthetic trajectories for each mechanism required approximately $22$ wall-clock hours distributed across $120$ CPU cores (executed as four parallel 30-core jobs, each provisioned with 240GB of RAM). Training of the full architecture was executed in roughly $3$ hours per training configuration utilizing $4$ CPU cores, 16GB of RAM, and a 10GB partition of an NVIDIA H100 GPU.

\newpage

\section{Simulation Details and Reaction Mechanisms}
\label{app:simulation_details}
This appendix provides the mathematical formulations for the candidate mechanisms evaluated in this study, alongside the parameters and process utilized for data simulation.

\subsection{Simulation Configuration}
All synthetic datasets were generated by numerically integrating the deterministic ODEs corresponding to each candidate CRN. All mechanism as well as the experimental settings to simulate from were taken from \citet{bures_organic_2023}. For every mechanism $M \in \mathcal{M}$, the kinetic constants were drawn randomly from independent log-uniform priors, i.e. $\log k_i \sim \mathcal{U}(\log 10^{-5}, \log 10^5)$. Each kinetic sample is used to produce $N_{\text{exp}} = 4$ concentration profiles. To induce varying kinetic regimes, the initial setups are configured as follows: The first three profiles are initialized with a standard substrate concentration of $[S]_0 = 1$ and product concentration of $[P]_0 = 0$. The catalyst loadings $[cat]_0$ for these first three profiles are uniformly sampled at random between 1 and 10 mol\%. The fourth profile represents a "same-excess" experiment: the substrate $[S]_0$ is initialized at a random value between 0.4 and 0.8, with the remaining mass balance added as initial product such that $[P]_0 = 1 - [S]_0$. The initial catalyst loading $[cat]_0$ for this fourth profile is randomly selected from the first three profiles. All other initial concentrations are set to zero. From each continuous kinetic profile, observations are evaluated at $T=20$ discrete concentration-time points randomly selected across the trajectory. These points are distorted using Gaussian noise model. The independent observation noise scales for the substrate ($\sigma_S$) and product ($\sigma_P$) are sampled from a log-normal prior: $\log \sigma \sim \mathcal{N}(-6, 1)$. To ensure stable network training and allow the architecture to generalize across varying kinetic regimes, both the temporal grids and the resulting species concentrations for each trajectory were min-max normalized to the range $[0, 1]$ prior to passing them to the summary network.   

Below we show the CRN for each mechanism and the corresponding systems of ODEs. Each node of the CRN corresponds to a chemical species. Arrows indicate which reactants (origin) lead to which products (end) for an intermediate reaction. The reaction rate constant dictating the propensity of each reaction is shown next to the corresponding arrow. Note that to ensure the generated trajectories accurately reflect their target structural hypotheses, the sampled rate constants for certain mechanisms must satisfy specific yield constraints. Solid arrows correspond to forward reactions, while dashed arrows to the reverse. In all equations, $[X]$ denotes the concentration of species $X$ at time $t$.   

\subsection{Reaction Mechanisms Considered}

\subsubsection{Mechanism 1}
The catalyst reversibly binds to the substrate to form an active complex, which then converts to the product.

\begin{center}
\begin{minipage}[c]{0.48\textwidth}
\centering
\begin{tikzpicture}[>=stealth, thick, scale=0.8, transform shape]
    \node (cat) at (0, 2) {cat};
    \node (catS) at (0, -2) {catS};
    \node (S) at (3, 2) {S};
    \node (P) at (-3, 2) {P};

    \coordinate (R1f) at (2.2, 0);
    \draw[-] (cat) to[bend left=15] (R1f);
    \draw[-] (S) to[bend left=15] (R1f);
    \draw[->] (R1f) to[bend left=15] node[right=0.1cm] {$k_1$} (catS);

    \coordinate (R1r) at (1.2, 0);
    \draw[dashed] (catS) to[bend right=15] (R1r) node[left=0.01cm] {$k_{-1}$};
    \draw[->, dashed] (R1r) to[bend right=15] (S);
    \draw[->, dashed] (R1r) to[bend right=15] (cat);

    \coordinate (R2f) at (-2.2, 0);
    \draw[-] (catS) to[bend left=15] node[left=0.1cm] {$k_2$} (R2f);
    \draw[->] (R2f) to[bend left=15] (cat);
    \draw[->] (R2f) to[bend right=15] (P);

    \coordinate (R2r) at (-1.2, 0);
    \draw[dashed] (cat) to[bend right=15] (R2r) node[right=0.1cm] {$k_{-2}$};
    \draw[dashed] (P) to[bend left=35] (R2r);
    \draw[->, dashed] (R2r) to[bend right=15] (catS);
\end{tikzpicture}
\end{minipage}
\hfill
\begin{minipage}[c]{0.48\textwidth}
\begin{equation*}
\label{eq:ode_m1}
\begin{aligned}
    \frac{d[S]}{dt} &= k_{-1}[catS] - k_1[S][cat] \\
    \frac{d[P]}{dt} &= k_2[catS] - k_{-2}[cat][P] \\
    \frac{d[cat]}{dt} &= (k_{-1} + k_2)[catS] - (k_1[S] + k_{-2}[P])[cat] \\
    \frac{d[catS]}{dt} &= (k_1[S] + k_{-2}[P])[cat] - (k_{-1} + k_2)[catS]
\end{aligned}
\end{equation*}
\end{minipage}
\end{center}

\subsubsection{Mechanism 2}
Mechanism 2 incorporates an off-cycle equilibrium where the active catalyst can reversibly form an inactive dimer ($cat_2$). It is defined by any set of kinetic constants such that a reaction run with $[S]_0 = 1$ and $[cat]_0=5$ mol\% produces a significant amount $cat_2$ (> 10\% of the total catalyst) during the first half of the reaction (between 20 and 50\% of the maximum conversion possible).

\begin{center}
\begin{minipage}[l]{0.48\textwidth}
\centering
\begin{tikzpicture}[>=stealth, thick, scale=0.8, transform shape]
    \node (cat) at (0, 2) {cat};
    \node (catS) at (0, -2) {catS};
    \node (cat2) at (0, 4.5) {cat$_2$};
    \node (S) at (3, 2) {S};
    \node (P) at (-3, 2) {P};

    \draw[->, dashed] (cat) to[bend right=20] node[right=0.1cm] {$k_{-3}$} (cat2);
    \draw[->] (cat2) to[bend right=20] node[left=0.1cm] {$k_3$} (cat);

    \coordinate (R1f) at (2.2, 0);
    \draw[-] (cat) to[bend left=15] (R1f);
    \draw[-] (S) to[bend left=15] (R1f);
    \draw[->] (R1f) to[bend left=15] node[right=0.1cm] {$k_1$} (catS);

    \coordinate (R1r) at (1.2, 0);
    \draw[dashed] (catS) to[bend right=15] (R1r) node[left=0.01cm] {$k_{-1}$};
    \draw[->, dashed] (R1r) to[bend right=15] (S);
    \draw[->, dashed] (R1r) to[bend right=15] (cat);

    \coordinate (R2f) at (-2.2, 0);
    \draw[-] (catS) to[bend left=15] node[left=0.1cm] {$k_2$} (R2f);
    \draw[->] (R2f) to[bend left=15] (cat);
    \draw[->] (R2f) to[bend right=15] (P);

    \coordinate (R2r) at (-1.2, 0);
    \draw[dashed] (cat) to[bend right=15] (R2r) node[right=0.1cm] {$k_{-2}$};
    \draw[dashed] (P) to[bend left=35] (R2r);
    \draw[->, dashed] (R2r) to[bend right=15] (catS);
\end{tikzpicture}
\end{minipage}
\hfill
\begin{minipage}[c]{0.5\textwidth}
\begin{equation*}
\label{eq:ode_m2}
\begin{aligned}     \frac{d[S]}{dt} &= k_{-1}[catS] - k_1[S][cat] \\     
\frac{d[P]}{dt} &= k_2[catS] - k_{-2}[cat][P] \\     
\frac{d[cat]}{dt} &= (k_{-1} + k_2)[catS] + 2k_3[cat_2] \\
& \quad - (k_1[S] + k_{-2}[P] + 2k_{-3}[cat])[cat] \\     \frac{d[catS]}{dt} &= (k_1[S] + k_{-2}[P])[cat] - (k_{-1} + k_2)[catS] \\     
\frac{d[cat_2]}{dt} &= k_{-3}[cat]^2 - k_3[cat_2] 
\end{aligned}
\end{equation*}
\end{minipage}
\end{center}

\subsubsection{Mechanism 3}
Mechanism 3 is characterized by an off-cycle resting state where the catalytic dimer binds with the substrate to form a $(cat)_2S$ complex, which is most significant during the initial stages of conversion. It is defined by any set of kinetic constants such that the reaction run $[S]_0 = 1$ and $[cat]_0=5$ mol\% has a significant amount of $cat$ (> 5\% of the total catalyst) at the beginning of the reaction (< 10\% of the maximum conversion possible).

\begin{center}
\begin{minipage}[c]{0.48\textwidth}
\centering
\begin{tikzpicture}[>=stealth, thick, scale=0.8, transform shape]
    \node (cat2) at (0, 2) {cat$_2$};
    \node (cat2S) at (0, -2) {(cat)$_2$S};
    \node (cat) at (0, 4.5) {cat};
    \node (S) at (3, 2) {S};
    \node (P) at (-3, 2) {P};

    \draw[->] (cat) to[bend left=20] node[right=0.1cm] {$k_3$} (cat2);
    \draw[->, dashed] (cat2) to[bend left=20] node[left=0.1cm] {$k_{-3}$} (cat);

    \coordinate (R1f) at (2.2, 0);
    \draw[-] (cat2) to[bend left=15] (R1f);
    \draw[-] (S) to[bend left=15] (R1f);
    \draw[->] (R1f) to[bend left=15] node[right=0.1cm] {$k_1$} (cat2S);

    \coordinate (R1r) at (1.2, 0);
    \draw[dashed] (cat2S) to[bend right=15] (R1r) node[left=0.01cm] {$k_{-1}$};
    \draw[->, dashed] (R1r) to[bend right=15] (S);
    \draw[->, dashed] (R1r) to[bend right=15] (cat2);

    \coordinate (R2f) at (-2.2, 0);
    \draw[-] (cat2S) to[bend left=15] node[left=0.1cm] {$k_2$} (R2f);
    \draw[->] (R2f) to[bend left=15] (cat2);
    \draw[->] (R2f) to[bend right=15] (P);

    \coordinate (R2r) at (-1.2, 0);
    \draw[dashed] (cat2) to[bend right=15] (R2r) node[right=0.1cm] {$k_{-2}$};
    \draw[dashed] (P) to[bend left=35] (R2r);
    \draw[->, dashed] (R2r) to[bend right=15] (cat2S);
\end{tikzpicture}
\end{minipage}
\hfill
\begin{minipage}[c]{0.48\textwidth}
\begin{equation*}
\label{eq:ode_m3}
\begin{aligned}
 \frac{d[S]}{dt} &= k_{-1}[(cat)_2S] - k_1[S][cat_2] \\
 \frac{d[P]}{dt} &= k_2[(cat)_2S] - k_{-2}[cat_2][P] \\     
 \frac{d[cat_2]}{dt} &= (k_{-1} + k_2)[(cat)_2S] + k_3[cat]^2 \\
 & \quad - (k_1[S] + k_{-2}[P] + k_{-3})[cat_2] \\     
 \frac{d[(cat)_2S]}{dt} &= (k_1[S] + k_{-2}[P])[cat_2] - (k_{-1} + k_2)[(cat)_2S] \\     
 \frac{d[cat]}{dt} &= 2k_{-3}[cat_2] - 2k_3[cat]^2
\end{aligned}
\end{equation*}
\end{minipage}
\end{center}

\subsubsection{Mechanism 5}
\label{app:M5}
Mechanism 5 evaluates the framework's scalability by introducing a pathway where the catalyst reversibly binds to the formed product, creating $catP$ complex. A second catalyst molecule is required to assist in the product release

\begin{center}
\begin{minipage}[c]{0.48\textwidth}
\centering
\begin{tikzpicture}[>=stealth, thick, scale=0.8, transform shape]
    \node (cat) at (0, 2) {cat};
    \node (catS) at (2.5, -2) {catS};
    \node (catP) at (-2.5, -2) {catP};
    \node (S) at (4, 2) {S};
    \node (P) at (-4, 2) {P};
    \node (cat_bottom) at (0, -4.5) {cat};

    \coordinate (R1f) at (2.2, 0.5);
    \draw[-] (cat) to[bend left=10] (R1f);
    \draw[-] (S) to[bend left=10] (R1f);
    \draw[->] (R1f) to[bend left=10] node[right=0.1cm] {$k_1$} (catS);

    \coordinate (R1r) at (1.2, 0.5);
    \draw[dashed] (catS) to[bend left=10] (R1r) node[left=0.01cm] {$k_{-1}$};
    \draw[->, dashed] (R1r) to[bend left=10] (cat);
    \draw[->, dashed] (R1r) to[bend right=10] (S);

    \coordinate (R3f) at (-2.2, 0.5);
    \draw[-] (catP) to[bend left=10] node[left=0.1cm] {$k_3$} (R3f);
    \draw[->] (R3f) to[bend left=10] (cat);
    \draw[->] (R3f) to[bend right=10] (P);

    \coordinate (R3r) at (-1.2, 0.5);
    \draw[dashed] (cat) to[bend left=10] (R3r) node[right=.1] {$k_{-3}$};
    \draw[dashed] (P) to[bend left=30] (R3r);
    \draw[->, dashed] (R3r) to[bend left=10] (catP);

    \coordinate (R2f) at (0, -3.2);
    \draw[-] (catS) to[bend left=10] (R2f);
    \draw[-] (cat_bottom) to[bend right=30] (R2f);
    \draw[->] (R2f) to[bend left=10] node[below=0.1cm] {$k_2$} (catP);
    \draw[->] (R2f) to[bend right=30] (cat_bottom);

    \coordinate (R2r) at (0, -0.5);
    \draw[dashed] (catP) to[bend right=15] (R2r);
    \draw[dashed] (cat) to[bend right=15] (R2r) node[below=0.2cm] {$k_{-2}$};
    \draw[->, dashed] (R2r) to[bend right=15] (catS);
    \draw[->, dashed] (R2r) to[bend left=15] (cat);
\end{tikzpicture}
\end{minipage}
\hfill
\begin{minipage}[c]{0.48\textwidth}
\begin{equation*}
\label{eq:ode_m5}
\begin{aligned}
\frac{d[S]}{dt} &= k_{-1}[catS] - k_1[S][cat] \\     
\frac{d[P]}{dt} &= k_3[catP] - k_{-3}[cat][P] \\     
\frac{d[cat]}{dt} &= k_{-1}[catS] + k_3[catP] - (k_1[S] + k_{-3}[P])[cat] \\     
\frac{d[catS]}{dt} &= (k_1[S] + k_{-2}[catP])[cat] - (k_{-1} + k_2[cat])[catS] \\     
\frac{d[catP]}{dt} &= k_2[catS][cat] + k_{-3}[P][cat] - (k_3 + k_{-2}[cat])[catP]
\end{aligned}
\end{equation*}
\end{minipage}
\end{center}

\clearpage

\section{Extended Simulation Results}
\label{app:extended_results}

Here we provide additional results supporting the main text. We include performance metrics for all the implementations described in our simulation study. To maintain clarity, full visual diagnostics are presented only for the highest-resolution setting (Setup 1: both chemical species, $T=20$). For the remaining regimes, inference performance is summarized via the evaluation metrics.

\subsection{SMC-ABC Results}
\label{app:results_abc}

To establish a performance baseline for our parameter inference and model selection framework, we employed an SMC-ABC algorithm. The distance metric was defined as the $L_2$ norm between the "experimental" trajectories and simulated trajectories generated from candidate sets of parameter values. The results presented here correspond to a representative case where the true mechanism underlying the observed synthetic data is $M_2$. We implemented the adaptive threshold SMC-ABC algorithm across the candidate models, initialized with $\epsilon_0 = 8.5$ and an adaptive threshold quantile of $\alpha = 0.55$. For this single observation, the algorithm required evaluating $928,605$ simulations to produce $3,000$ samples for each model across $6$ generations.

\begin{figure}[ht]
    \centering
    \includegraphics[width=0.9\linewidth]{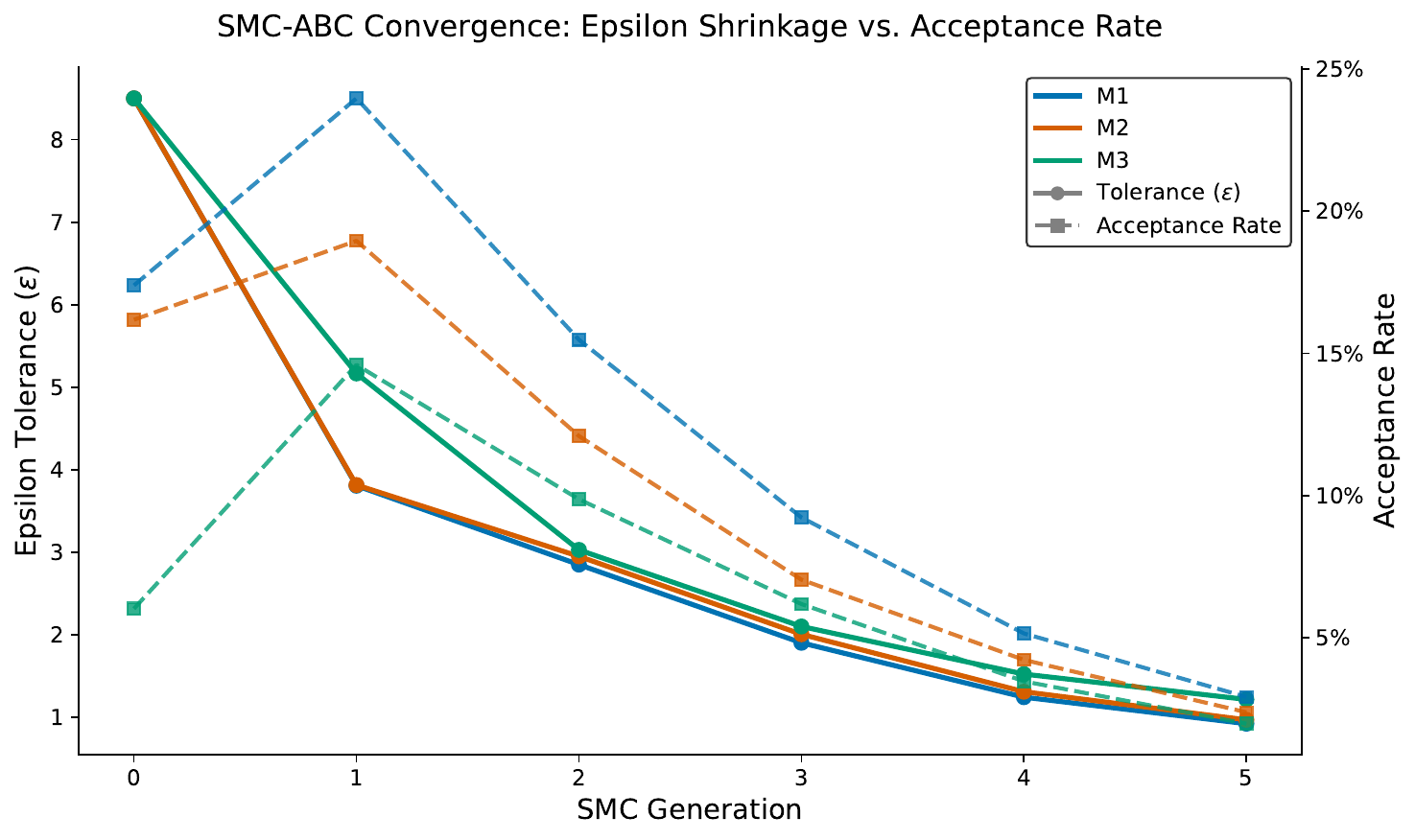}
    \caption{Convergence diagnostics for SMC-ABC inference. The solid lines (left axis) track the shrinkage of the discrepancy threshold ($\epsilon$) across generations for candidate mechanisms M1, M2, and M3. The dashed lines (right axis) denote the corresponding acceptance rates required to maintain the target population of $N=3,000$ particles per generation. As the posterior approximation refines and the $\epsilon$ tolerance tightens, the acceptance rate exhibits a rapid decline, reflecting the increasing computational cost of proposing valid kinetic parameters from the transition kernel.}
    \label{fig:smc_abc_convergence}
\end{figure}

Figure \ref{fig:smc_abc_convergence} highlights the computational inefficiencies of SMC-ABC in the context of our problem. SMC-ABC requires new simulations every time inference is performed on new data. Inference is costly as several simulations must be produced for one parameter sample to be accepted. The acceptance rate curves in the figure illustrate the exponential increase in required simulations as the tolerance threshold $\epsilon$ shrinks. From the first to the sixth generation, the acceptance rates dropped from $17.4\%$, $16.2\%$, and $6.1\%$ down to $2.9\%$, $2.4\%$, and $2.0\%$ for $M_1$, $M_2$, and $M_3$, respectively. This also demonstrates that as model complexity increases, acceptance rates become substantially lower, even during initial generations.

\begin{figure}[h]
    \centering
    \includegraphics[width=\linewidth, trim={0cm 1.2cm 0cm 1cm}, clip]{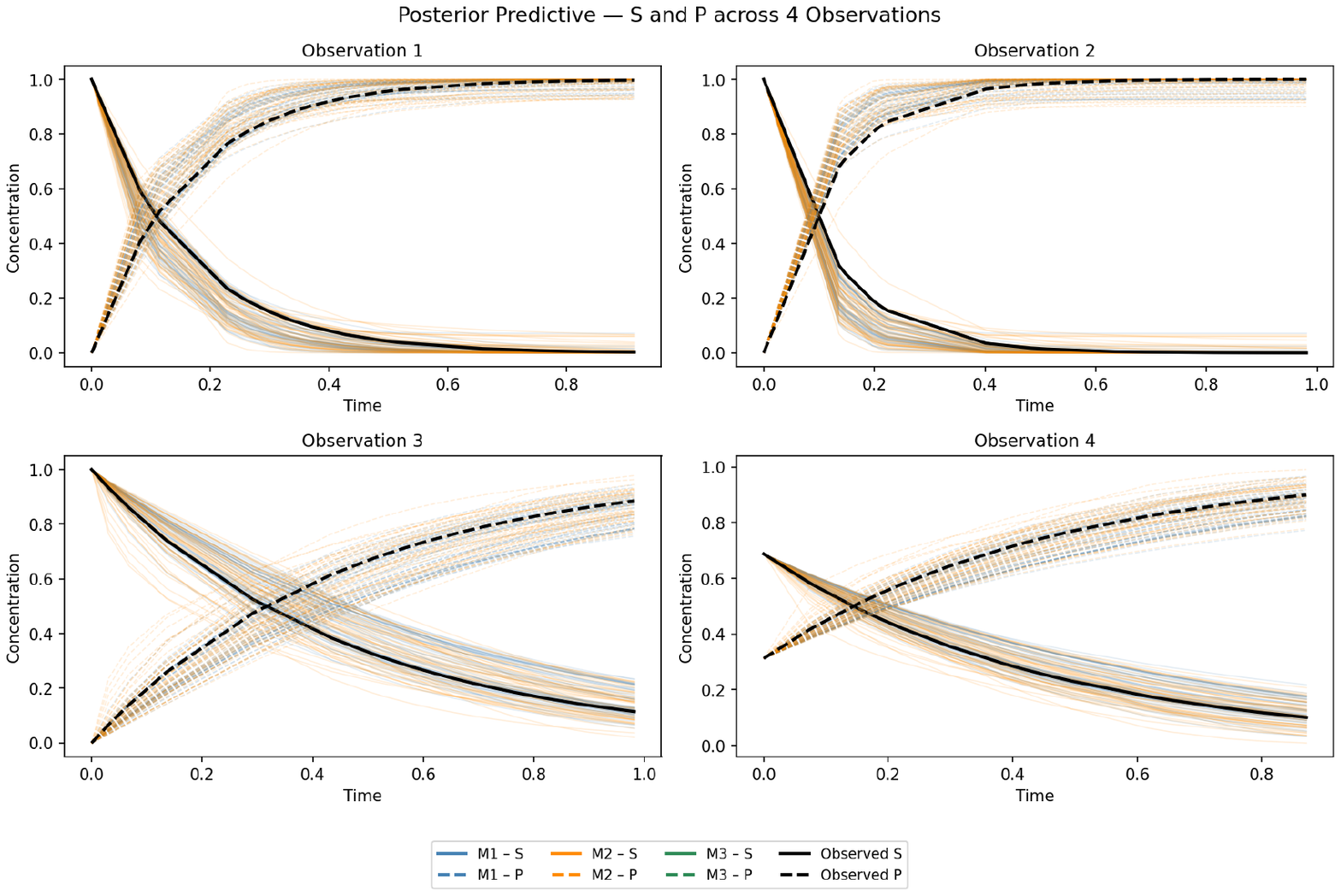}
    \caption{PPC for the four experimental observations, where SMC-ABC was implemented. Black lines denote the true observed synthetic data, with solid and dashed lines representing the substrate (S) and product (P) concentrations, respectively. Faint colored lines depict the 100 inferred trajectories generated by drawing from the network's estimated posterior.}
    \label{fig:smc_abc_ppc}
\end{figure}

For the case presented here, our implementation of SMC-ABC failed to identify the ground-truth mechanism. The highest estimated posterior probability belonged to $M_1$ ($53.1\%$), as opposed to the true mechanism $M_2$ ($46.8\%$). Figure \ref{fig:smc_abc_ppc} presents a Posterior Predictive Check (PPC) for this case. To generate this PPC, a candidate mechanism is first sampled according to the estimated posterior probabilities, followed by sampling kinetic parameters from that model's estimated posterior. These inferred parameters are then passed through the ODE simulator to generate the 100 inferred trajectories. The resulting PPC shows considerable agreement between the observed data and the inferred trajectories. This outcome highlights the overlap within the candidate space: under certain parameter regimes, distinct CRNs can produce very similar macroscopic concentration profiles. 

\subsection{Amortized Neural Inference Results (3 Models)}
\label{app:results_3model}

In our primary study considering $\mathcal{M} = \{M_1, M_2, M_3\}$, the amortized framework demonstrates both high classification accuracy and well-calibrated parameter estimation. This section details the performance metrics and presents relevant figures resulting from implementing our approach under four distinct observational setups: (1) both substrate and product observed across all $T=20$ discrete time points; (2) both species observed at $T=10$ random time points; (3) only the product observed across all $T=20$ time points; and (4) only the product observed at $T=10$ random time points.

In all cases, performance was evaluated using a holdout test set of $3,000$ synthetic observations ($1,000$ per true mechanism). We used two trained networks to handle all four setups. The first network was trained on trajectories observing both species, enabling inference for setups (1) and (2). The second network was trained observing only the product, for inference under setups (3) and (4). Both networks were trained on trajectories with a randomized number of observed time points (uniformly sampled between $5$ and $20$). During the evaluation for setups (2) and (4), the $10$ available time points were a random subset from the full $20$-point temporal grid.

Table \ref{tab:classification_3model} details the confusion matrix and corresponding classification metrics for all setups of observation, demonstrating the network's capacity to distinguish the underlying CRN structures. For model selection, the predicted mechanism for each test trajectory is chosen to be the one with the highest posterior probability (i.e., the maximum a posteriori estimate). The metrics shown compare this predicted mechanism with the true mechanism in each test case.
\begin{table}[ht]
\centering
\caption{Classification Performance. Results for the three-model setting ($\mathcal{M} = \{M_1, M_2, M_3\}$) across all four observational setups. The left pane displays the confusion matrix across the holdout test set, and the right pane details the corresponding performance metrics.}
\label{tab:classification_3model}
\begin{tabular}{@{}llccccccc@{}}
\toprule
& & \multicolumn{3}{c}{\textbf{Predicted Model}} & & \multicolumn{3}{c}{\textbf{Performance Metrics}} \\
\cmidrule{3-5} \cmidrule{7-9}
\textbf{Setup} & \textbf{Truth} & $M_1$ & $M_2$ & $M_3$ & & Precision & Recall & $F_1$-score \\
\midrule
(1) S \& P, $T=20$ & $M_1$ & 954 & 9 & 37 & & 0.93 & 0.95 & 0.94 \\
                   & $M_2$ & 18 & 976 & 6 & & 0.99 & 0.98 & 0.98 \\
                   & $M_3$ & 51 & 5 & 944 & & 0.96 & 0.94 & 0.95 \\
\midrule
(2) S \& P, $T=10$ & $M_1$ & 936 & 17 & 47 & & 0.92 & 0.94 & 0.93 \\
                   & $M_2$ & 19 & 965 & 16 & & 0.98 & 0.96 & 0.97 \\
                   & $M_3$ & 64 & 7 & 929 & & 0.94 & 0.93 & 0.93 \\
\midrule
(3) P only, $T=20$ & $M_1$ & 943 & 14 & 43 & & 0.88 & 0.94 & 0.91 \\
                   & $M_2$ & 23 & 971 & 6 & & 0.98 & 0.97 & 0.97 \\
                   & $M_3$ & 103 & 6 & 891 & & 0.95 & 0.89 & 0.92 \\
\midrule
(4) P only, $T=10$ & $M_1$ & 901 & 28 & 71 & & 0.88 & 0.90 & 0.89 \\
                   & $M_2$ & 28 & 951 & 21 & & 0.96 & 0.95 & 0.95 \\
                   & $M_3$ & 99 & 12 & 889 & & 0.91 & 0.89 & 0.90 \\
\bottomrule
\end{tabular}
\end{table}

To evaluate the calibration of the estimated posterior probabilities, we computed the Expected Calibration Error (ECE) and evaluated the corresponding reliability diagrams. A perfectly calibrated summary network would produce an ECE of zero and a reliability curve along the identity line, indicating that when the network assigns a $p$ probability to a specific mechanism, that prediction turns out to be correct in exactly $(100*p)\%$ of the cases.

\begin{figure*}[ht]
    \centering
    
    
    \begin{minipage}[t]{0.53\textwidth}
    \vspace{0pt}
        \centering
        \includegraphics[width=\linewidth]{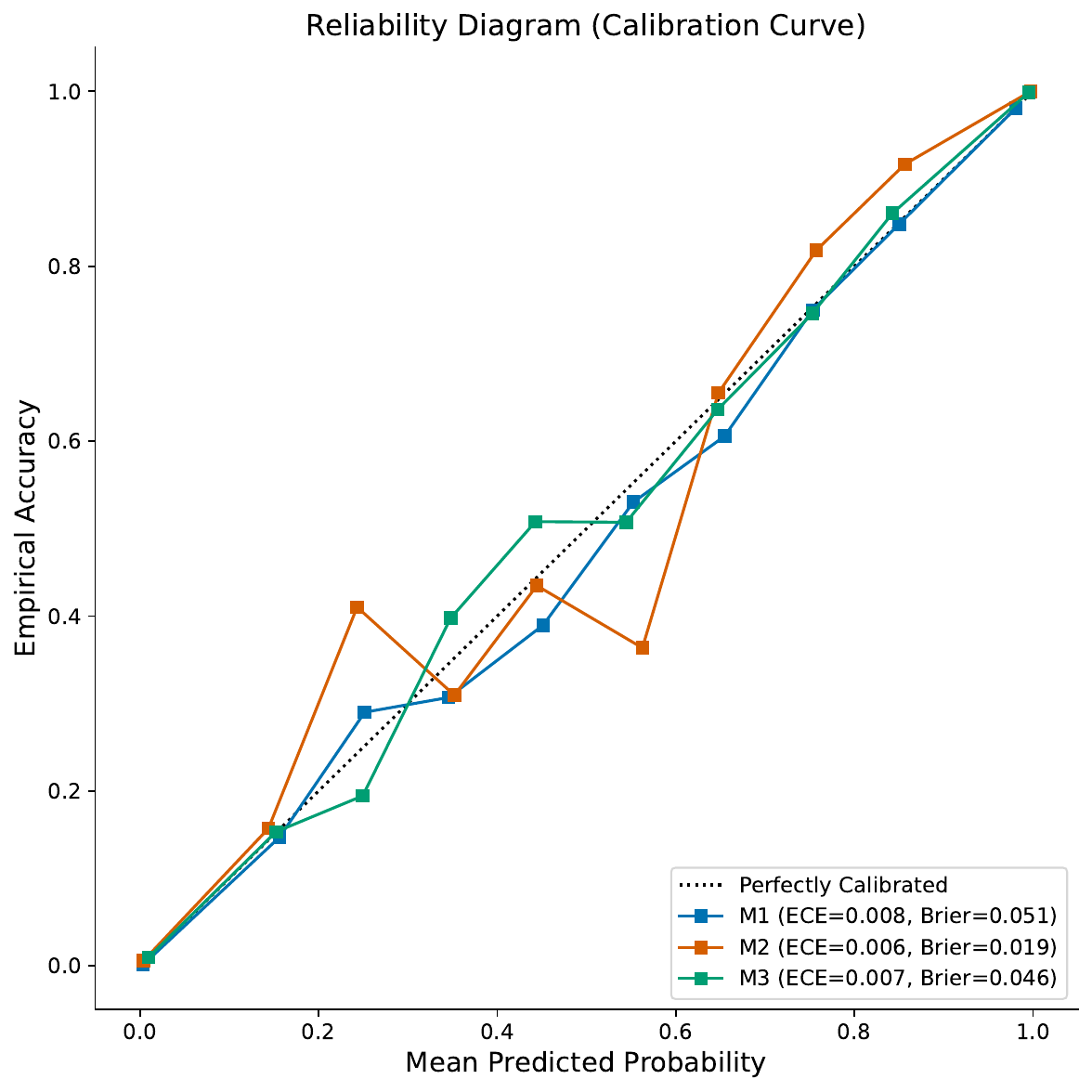} 
        \makeatletter\def\@captype{figure}\makeatother 
        \caption{Reliability diagram evaluating the calibration of the posterior probabilities across the 3,000 test trajectories (only $P$ is observed, $T=10$). Predictions are grouped into 10 uniformly spaced probability bins. The dashed diagonal represents perfect calibration, where the estimated probability exactly matches empirical accuracy.}
        \label{fig:reliability_3model}
    \end{minipage} 
    \hfill
    \begin{minipage}[t]{0.46\textwidth}
    \vspace{85pt}
        \centering
        \makeatletter\def\@captype{table}\makeatother 
        \caption{Expected Calibration Error. ECE for posterior probabilities across the four observational setups. Lower values indicate better alignment between estimated probabilities and testing accuracy.}
        \label{tab:ece_3model}
        \vspace{0.15cm} 
        \begin{tabular}{@{}lccc@{}}
        \toprule
        & \multicolumn{3}{c}{\textbf{ECE}} \\
        \cmidrule{2-4}
        \textbf{Setup} & $M_1$ & $M_2$ & $M_3$ \\
        \midrule
        (1) S \& P, $T=20$ & 0.0060 & 0.0056 & 0.0059 \\
        (2) S \& P, $T=10$ & 0.0083 & 0.0064 & 0.0107 \\
        (3) P only, $T=20$ & 0.0124 & 0.0052 & 0.0094 \\
        (4) P only, $T=10$ & 0.0078 & 0.0064 & 0.0071 \\
        \bottomrule
        \end{tabular}
    \end{minipage}

\end{figure*}

All reliability diagrams were constructed by splitting the network's outputted posterior probabilities across all test cases into $10$ equally spaced bins across the $[0, 1]$ interval. For each bin, we computed the average posterior probability and the empirical accuracy (the fraction of cases from the bin that actually correspond to the mechanism evaluated). The ECE was then calculated as the weighted average of the absolute differences between the mean probability and the empirical accuracy across all bins (weighted by the number of cases per bin). Calibration results for each observational setup are presented in Table \ref{tab:ece_3model}. As the observational data becomes sparser, the ECE remains low, fluctuating around $1\%$. This suggests that when missing sufficient data to uniquely identify the true mechanism, the network appropriately scales back its estimated probabilities, correctly quantifying the uncertainty about the true mechanism. The reliability diagram for the main setup is not presented here as it is present in the main text (Figure \ref{fig:calibration}). Figure shows \ref{fig:reliability_3model} the diagram that corresponds to observational setup (4). Even in this case, with only one of the species available and observing only $10$ time steps, the posterior probabilities produced by the model network are well-calibrated. 

Beyond model selection, the network facilitates parameter estimation by generating samples from the models' estimated posterior distributions. Figure \ref{fig:posteriors_1} illustrates the estimated marginal posteriors ($2,000$ samples) for a representative test trajectory generated by mechanism $M_3$ (main observational setup). For this test case, the network accurately identified the underlying structural pathway, assigning it a posterior probability approaching $1$. To evaluate the reliability of these estimates across the entire test set, we calculate the coverage of the $95\%$ credible intervals---tracking the percentage of cases where the true parameter value falls within this bound. Table \ref{tab:master_coverage_3model} contains the coverage results for all four settings.  

\begin{table*}[h!]
\centering
\caption{Empirical Coverage for Three Models. Coverage (\%) for the $95\%$ credible intervals for the kinetic parameters and observation noise across all four observational setups. Values close to $95\%$ indicate well-calibrated marginal posteriors.}
\label{tab:master_coverage_3model}
\begin{tabular}{@{}lcccccccccccc@{}}
\toprule
& \multicolumn{3}{c}{\textbf{(1) S \& P, $T=20$}} & \multicolumn{3}{c}{\textbf{(2) S \& P, $T=10$}} & \multicolumn{3}{c}{\textbf{(3) P only, $T=20$}} & \multicolumn{3}{c}{\textbf{(4) P only, $T=10$}} \\
\cmidrule(lr){2-4} \cmidrule(lr){5-7} \cmidrule(lr){8-10} \cmidrule(lr){11-13}
\textbf{Param} & $M_1$ & $M_2$ & $M_3$ & $M_1$ & $M_2$ & $M_3$ & $M_1$ & $M_2$ & $M_3$ & $M_1$ & $M_2$ & $M_3$ \\
\midrule
$k_1$      & 96.2 & 96.2 & 96.2 & 95.0 & 96.2 & 96.0 & 97.7 & 94.9 & 98.0 & 95.9 & 92.6 & 97.4 \\
$k_{-1}$   & 96.4 & 95.2 & 96.0 & 97.0 & 94.9 & 96.4 & 95.8 & 94.4 & 94.9 & 96.5 & 94.8 & 94.0 \\
$k_2$      & 98.1 & 97.3 & 95.1 & 97.5 & 94.3 & 95.4 & 97.8 & 95.2 & 93.9 & 96.5 & 94.6 & 93.9 \\
$k_{-2}$   & 97.0 & 94.8 & 96.8 & 97.7 & 94.8 & 97.3 & 97.0 & 95.4 & 94.3 & 96.6 & 94.6 & 94.8 \\
$k_3$      & --   & 93.0 & 95.2 & --   & 89.5 & 93.1 & --   & 97.7 & 94.0 & --   & 96.5 & 93.4 \\
$k_{-3}$   & --   & 96.2 & 96.0 & --   & 94.0 & 96.1 & --   & 95.9 & 95.2 & --   & 94.6 & 96.1 \\
\midrule
$\sigma_S$ & 95.1 & 93.0 & 95.8 & 97.4 & 97.0 & 97.1 & --   & --   & --   & --   & --   & --   \\
$\sigma_P$ & 95.1 & 93.0 & 94.3 & 97.5 & 97.2 & 96.9 & 95.9 & 97.3 & 96.6 & 93.7 & 94.0 & 95.0 \\
\bottomrule
\end{tabular}
\end{table*}

\begin{figure}[h!]
    \centering
    \includegraphics[width=\linewidth]{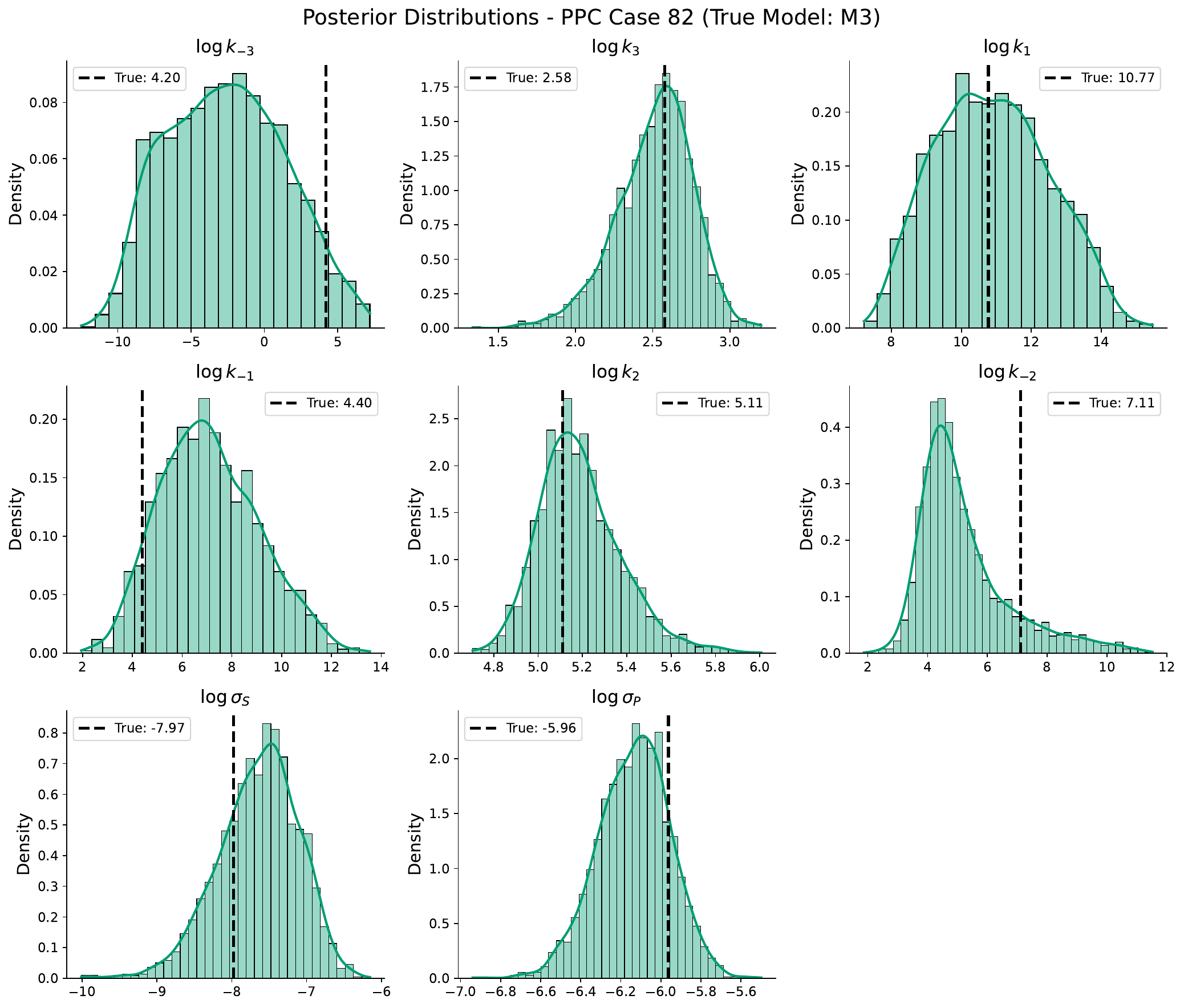}
    \caption{Estimated posterior distributions, using $2,000$ samples, for all parameters from a test case where the true mechanism was $M_3$. The true value (in the log-space) is marked with black dashed line.}
    \label{fig:posteriors_1}
\end{figure}

\begin{figure}[h!]
    \centering
    \includegraphics[width=.7\linewidth]{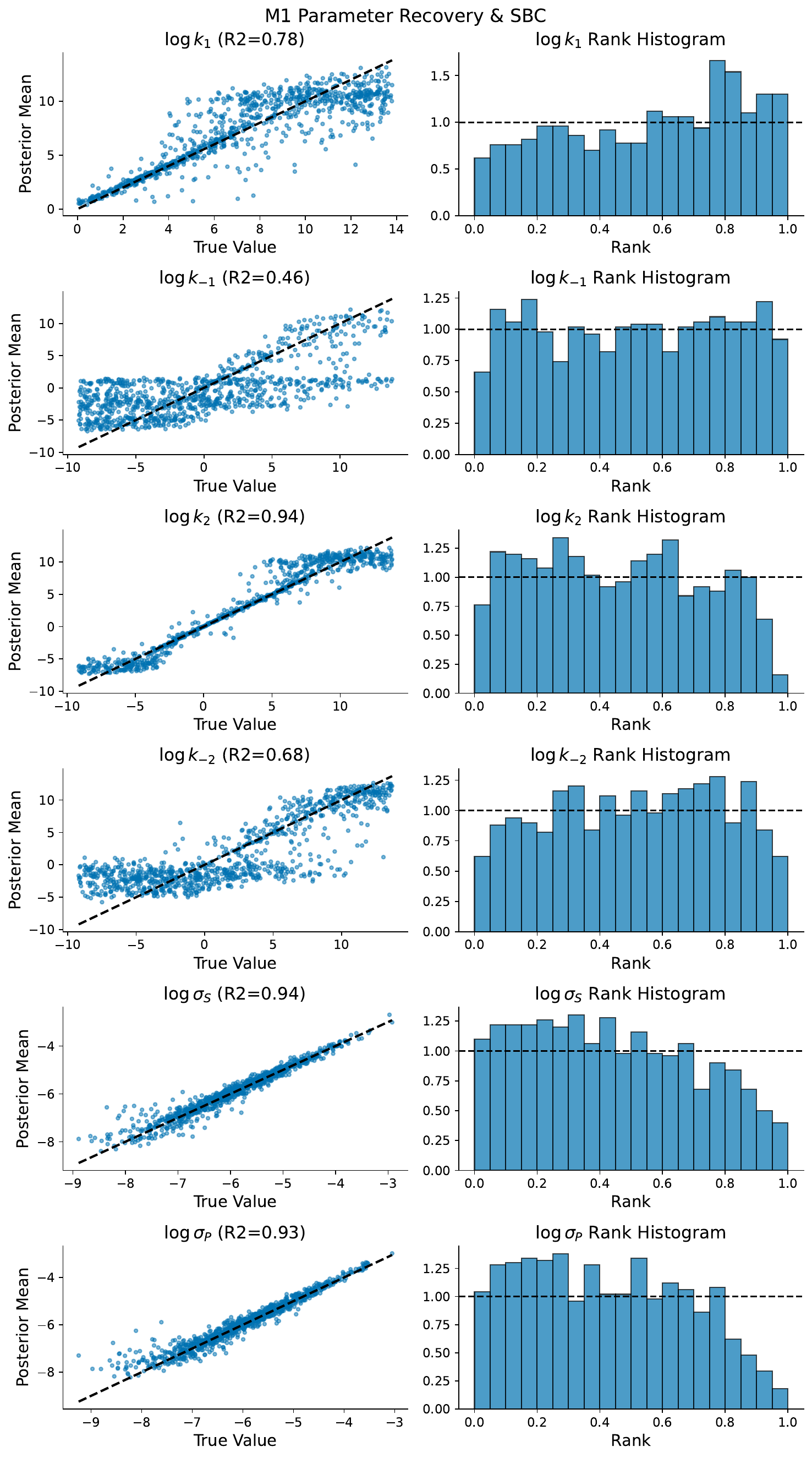}
    \caption{Inference calibration across all testing cases from $M_1$. Left Panel: Posterior mean estimates vs. the true values. Right Panel: Rank histograms for all parameters (normalized to $(0,1)$). The dashed black line indicated the level for uniform density ($1$).}
    \label{fig:rank_M1_1}
\end{figure}

\begin{figure}[h!]
    \centering
    \includegraphics[width=0.52\linewidth]{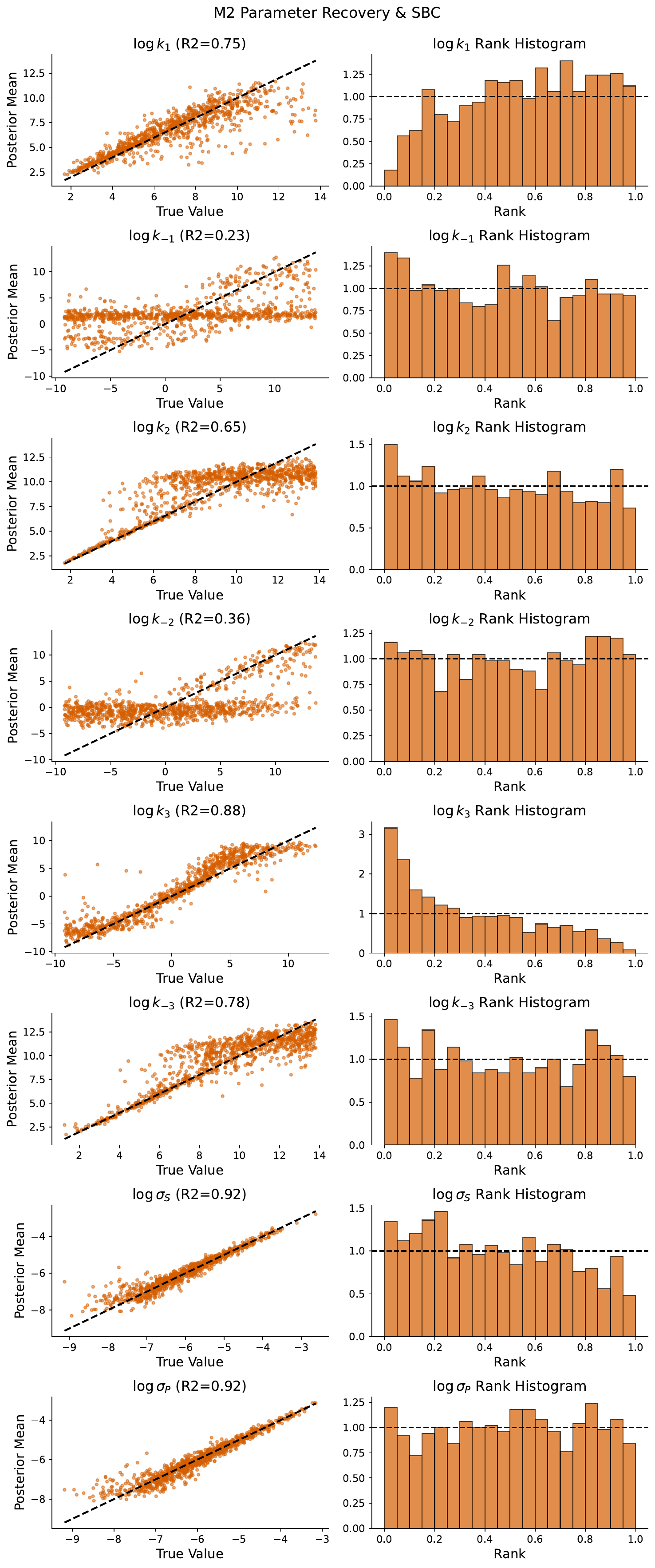}
    \caption{Inference calibration across all testing cases from $M_2$. Left Panel: Posterior mean estimates vs. the true values. Right Panel: Rank histograms for all parameters (normalized to $(0,1)$). The dashed black line indicated the level for uniform density ($1$).}
    \label{fig:rank_M2_1}
\end{figure}

\begin{figure}[h!]
    \centering
    \includegraphics[width=0.52\linewidth]{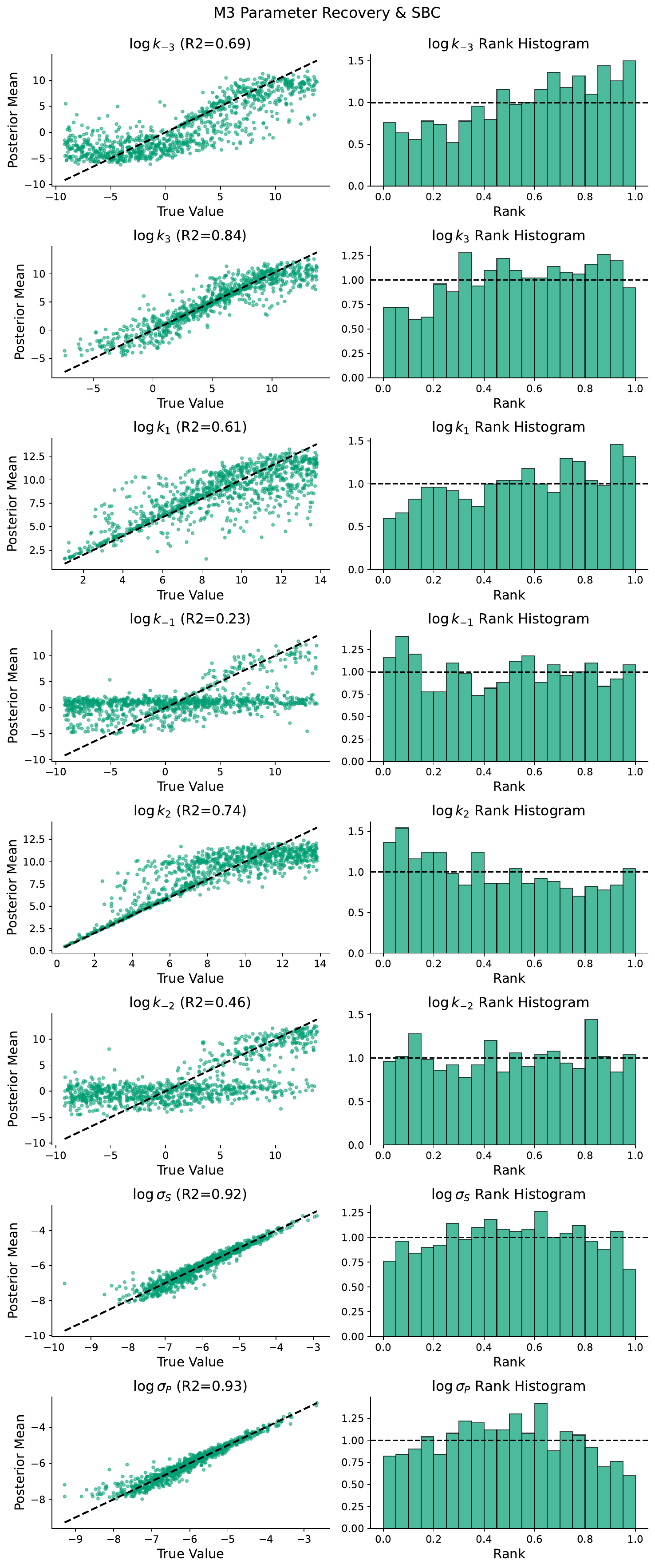}
    \caption{Inference calibration across all testing cases from $M_3$. Left Panel: Posterior mean estimates vs. the true values. Right Panel: Rank histograms for all parameters (normalized to $(0,1)$). The dashed black line indicated the level for uniform density ($1$).}
    \label{fig:rank_M3_1}
\end{figure}

Further, to assess the calibration of the entire posterior distributions, we rely on Simulation-Based Calibration (SBC) rank histograms. To construct the SBC rank histograms, we used the $2,000$ posterior samples for each of the $1,000$ test trajectories per mechanism. In each case, the true parameter value was ranked among the corresponding posterior samples, resulting in an integer rank between $0$ and $2,000$ (normalized to $1$). The histograms visualize the distribution of these empirical ranks across the entire test set; posterior calibration would produce a uniform distribution across all possible ranks. Figures \ref{fig:rank_M1_1}, \ref{fig:rank_M2_1} and \ref{fig:rank_M3_1} show the rank histograms (right pane) for all parameters of $M_1, M_2,$ and $M_3$ respectively. They also include a comparison between the posterior mean estimate of the parameters and the true value across all testing observations. As the plots show, histograms are mostly uniform, even when the parameters seem to be weakly identifiable (for example $k_1$). This suggests that the network correctly expresses uncertainty about the true parameter values. 

\begin{figure}[h!]
    \centering
    \includegraphics[width=\linewidth]{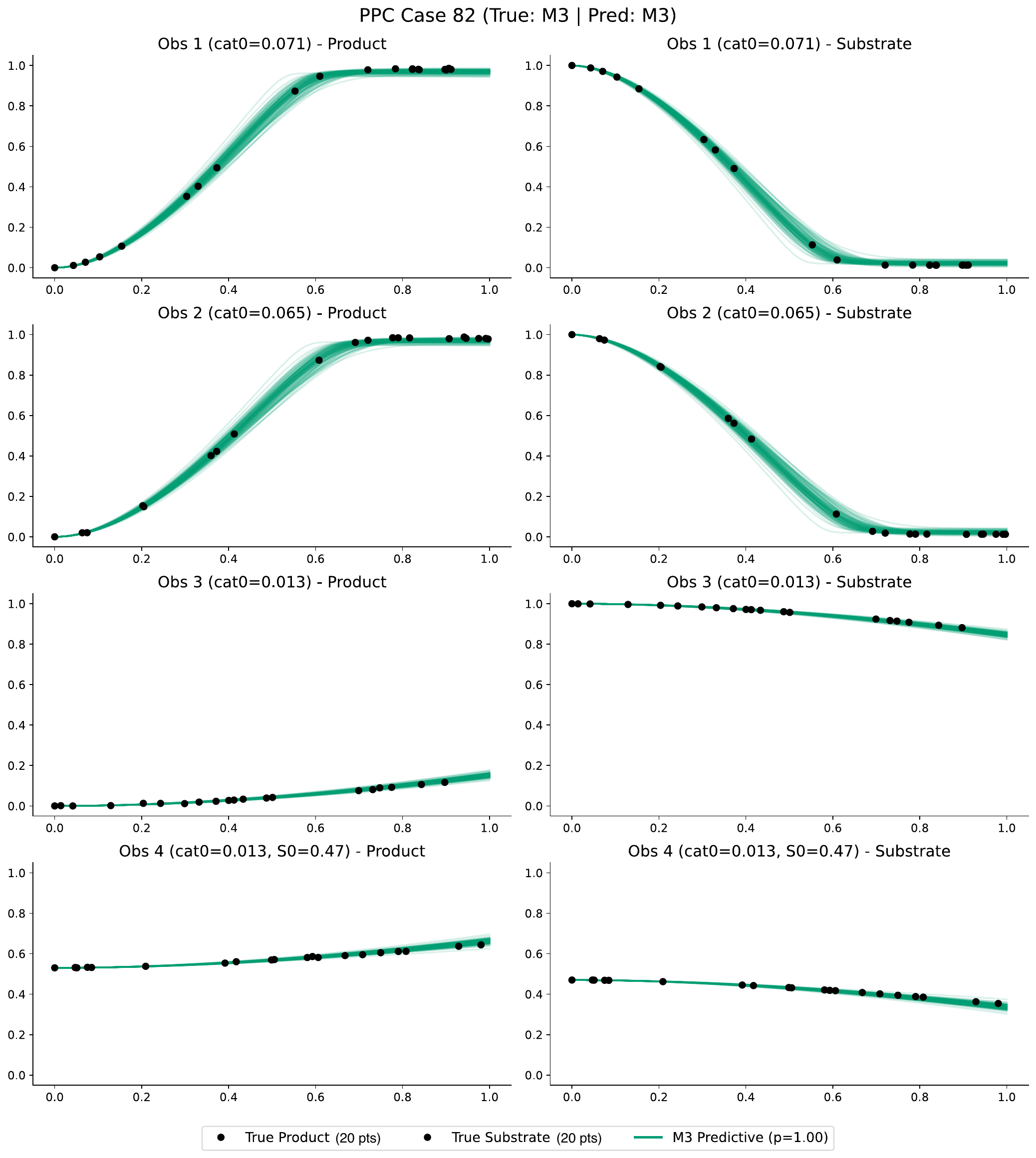}
    \caption{PPC for a test trajectory generated by mechanism $M_3$ under the high-resolution setup (1). Black dots denote the $20$ observations for both species across four distinct initial conditions. The solid lines represent $100$ posterior predictive trajectories sampled from the network's inferred joint posterior.}
    \label{fig:ppc_1}
\end{figure}

\begin{figure}[h!]
    \centering
    \includegraphics[width=\linewidth]{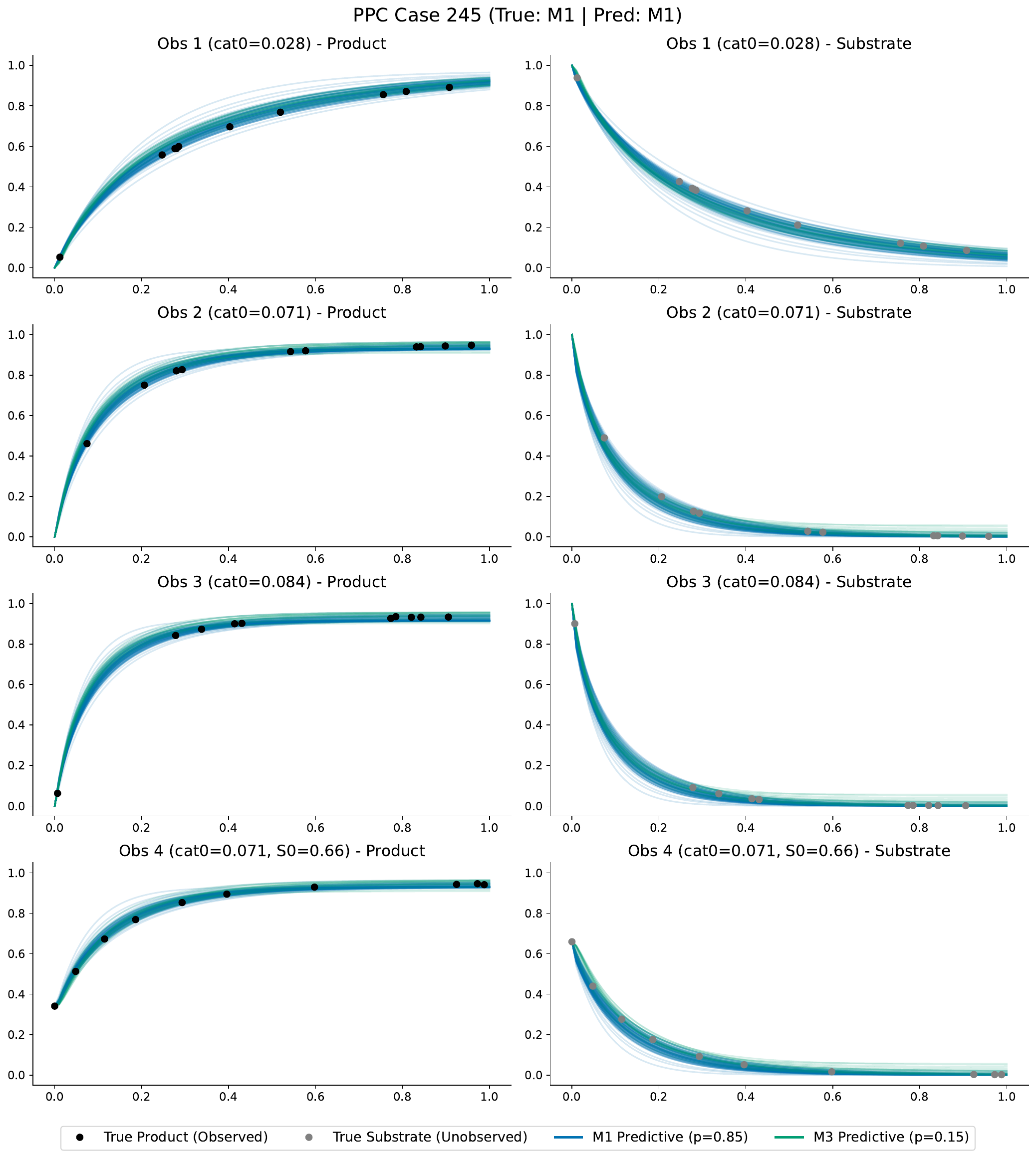}
    \caption{PPC for a test trajectory generated by mechanism $M_1$ under the low-resolution setup (4). Black dots denote the $10$ observations for the product across four distinct initial conditions. The gray dots indicate the corresponding substrate, which was not observed by the model. The solid lines represent $100$ posterior predictive trajectories sampled from the network's inferred joint posterior.}
    \label{fig:ppc_2}
\end{figure}

Lastly, we conduct Posterior Predictive Checks (PPCs) by simulating data ($100$ samples) from the inferred parameter distributions to evaluate how well our approach models the true concentration profiles. Figure \ref{fig:ppc_1} presents the PPC corresponding to the same test trajectory as the posterior distributions from Figure \ref{fig:posteriors_1} (generated by mechanism $M_3$ under the setup (1): both species observed, $T=20$). The network correctly identifies the CRN structure with absolute confidence and the predictive samples perfectly track the ground-truth trajectories for both the substrate and the product across all four experimental conditions. In contrast, Figure \ref{fig:ppc_2} illustrates the network's predictive fit under the sparse observational constraints of setup (4) (only the product observed, $T=10$). For this trajectory generated by mechanism $M_1$, the network successfully identifies the true mechanism but presents some model uncertainty, assigning a  posterior probability of $p_1 = 0.85$ to $M_1$ and allocating the remaining $p_3 = 0.15$ probability to mechanism $M_3$. The resulting PPC shows this uncertainty: predictive samples are drawn from both candidate models in accordance with their probabilities. Despite the absence of substrate data and a restricted $10$-point temporal resolution, the network not only encompasses the observed product data but successfully reconstructs the unobserved substrate dynamics with high accuracy. The slightly wider predictive variance cleanly reflects the network's uncertainty regarding the missing kinetic information.

\clearpage

\subsection{Scalability Results (4 Models)}
\label{app:results_4model}

To evaluate the scalability of our neural posterior framework, the model space was expanded to include Mechanism 5 ($\mathcal{M} = \{M_1, M_2, M_3, M_5\}$). Table \ref{tab:classification_4model} details the classification performance over the $4,000$-sample testing set under the high-resolution setting. While the overall classification accuracy remains high at $86\%$, the confusion matrix reveals the structural uncertainty introduced by the inclusion of $M_5$. Mechanisms $M_2$ and $M_3$ remain highly distinguishable, maintaining $F_1$-scores of $0.98$ and $0.93$ respectively. However, some confusion emerges between $M_1$ and $M_5$: $186$ $M_1$ trajectories are classified as $M_5$, and $190$ $M_5$ trajectories are classified as $M_1$. 

\begin{table}[h!]
\centering
\caption{Classification Performance for Four Models. Results for the expanded four-model setting ($\mathcal{M} = \{M_1, M_2, M_3, M_5\}$) under the main observational setup (both species observed, $T=20$). The left pane displays the confusion matrix across the holdout test set, and the right pane details the corresponding performance metrics.}
\label{tab:classification_4model}
\begin{tabular}{@{}lccccccccc@{}}
\toprule
& \multicolumn{4}{c}{\textbf{Predicted Model}} & & \multicolumn{3}{c}{\textbf{Performance Metrics}} \\
\cmidrule{2-5} \cmidrule{7-9}
\textbf{Truth} & $M_1$ & $M_2$ & $M_3$ & $M_5$ & & Precision & Recall & $F_1$-score \\
\midrule
$M_1$ & 789 & 6 & 19 & 186 & & 0.75 & 0.79 & 0.77 \\
$M_2$ & 17 & 977 & 6 & 0 & & 0.98 & 0.98 & 0.98 \\
$M_3$ & 54 & 9 & 928 & 9 & & 0.93 & 0.93 & 0.93 \\
$M_5$ & 190 & 4 & 43 & 763 & & 0.80 & 0.76 & 0.78 \\
\bottomrule
\end{tabular}
\end{table}

\begin{figure*}[h!]
    \centering
    
    \begin{minipage}[t]{0.53\textwidth}
        \vspace{0pt} 
        \centering
        \includegraphics[width=\linewidth, trim=1.1cm 4.2cm .7cm 1.3cm, clip]{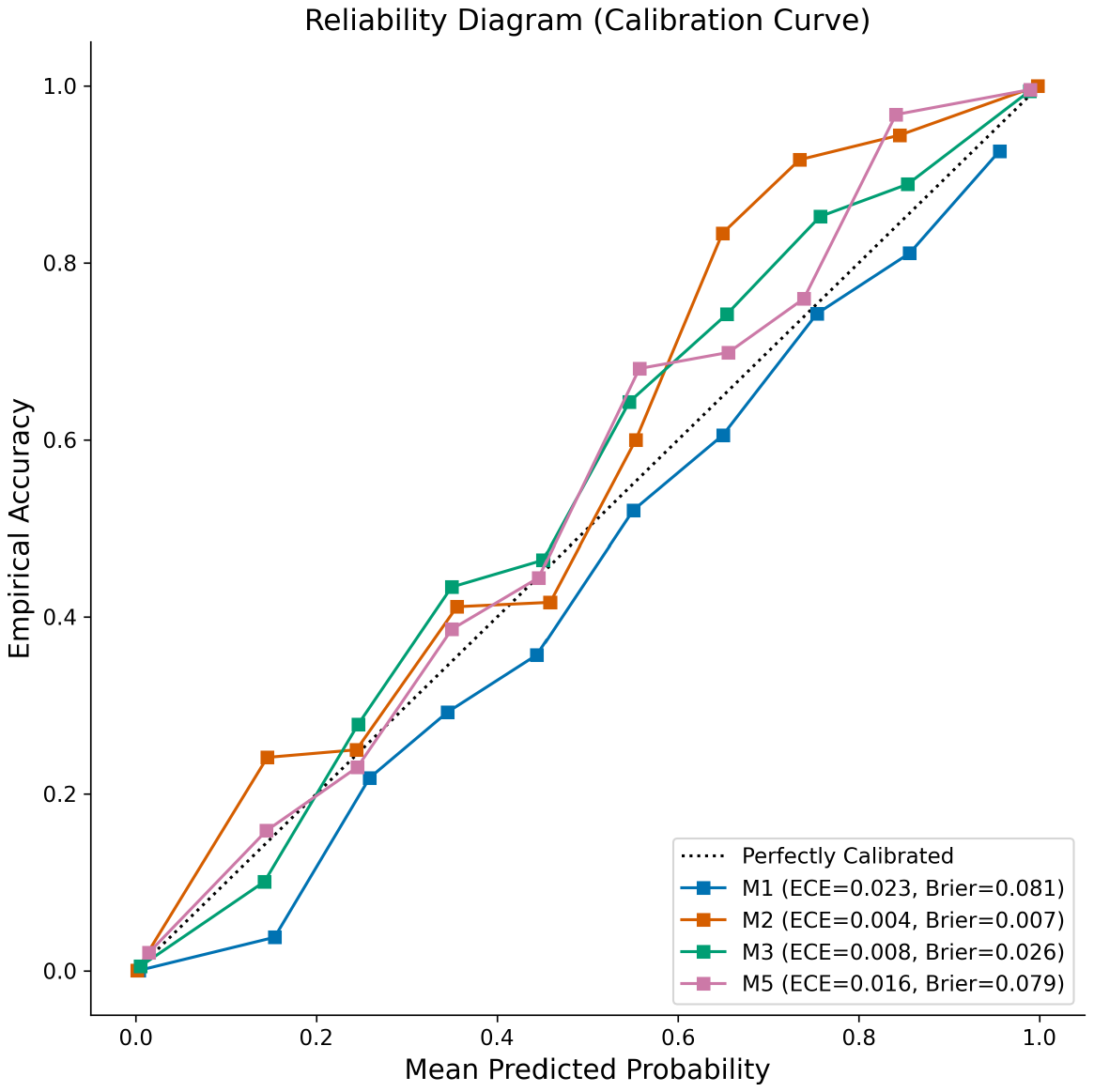} 
        \makeatletter\def\@captype{figure}\makeatother 
        \caption{Reliability diagram evaluating the calibration of the posterior probabilities across the 4,000 test trajectories ($P$ and $S$ observed, $T=20$). Predictions are grouped into 10 uniformly spaced probability bins. The dashed diagonal represents perfect calibration, where the estimated probability exactly matches empirical accuracy.}
        \label{fig:RD_4M}
    \end{minipage}\hfill
    \begin{minipage}[t]{0.46\textwidth}
        \vspace{85pt} 
        \centering
        \makeatletter\def\@captype{table}\makeatother 
        \caption{Empirical Coverage for Four Models. Coverage (\%) of the $95\%$ posterior credible intervals for the kinetic parameters and observation noise under the expanded four-model space ($\mathcal{M} = \{M_1, M_2, M_3, M_5\}$). Both substrate and product observed, $T=20$}.
        \label{tab:coverage_4model}
        \vspace{0.15cm} 
        \begin{tabular}{@{}lcccc@{}}
        \toprule
        \textbf{Parameter} & $M_1$ & $M_2$ & $M_3$ & $M_5$ \\
        \midrule
        $k_1$      & 97.1 & 94.7 & 94.9 & 97.8 \\
        $k_{-1}$   & 96.4 & 91.8 & 94.0 & 96.9 \\
        $k_2$      & 96.2 & 94.3 & 96.0 & 95.3 \\
        $k_{-2}$   & 95.3 & 93.4 & 96.9 & 96.2 \\
        $k_3$      & --   & 94.2 & 96.0 & 96.8 \\
        $k_{-3}$   & --   & 94.8 & 96.1 & 95.4 \\
        \midrule
        $\sigma_S$ & 96.4 & 95.4 & 95.8 & 96.6 \\
        $\sigma_P$ & 95.3 & 94.6 & 96.4 & 96.2 \\
        \bottomrule
        \end{tabular}
    \end{minipage}
    
\end{figure*}

\begin{figure}[h!]
    \centering
    \includegraphics[width=0.84\linewidth]{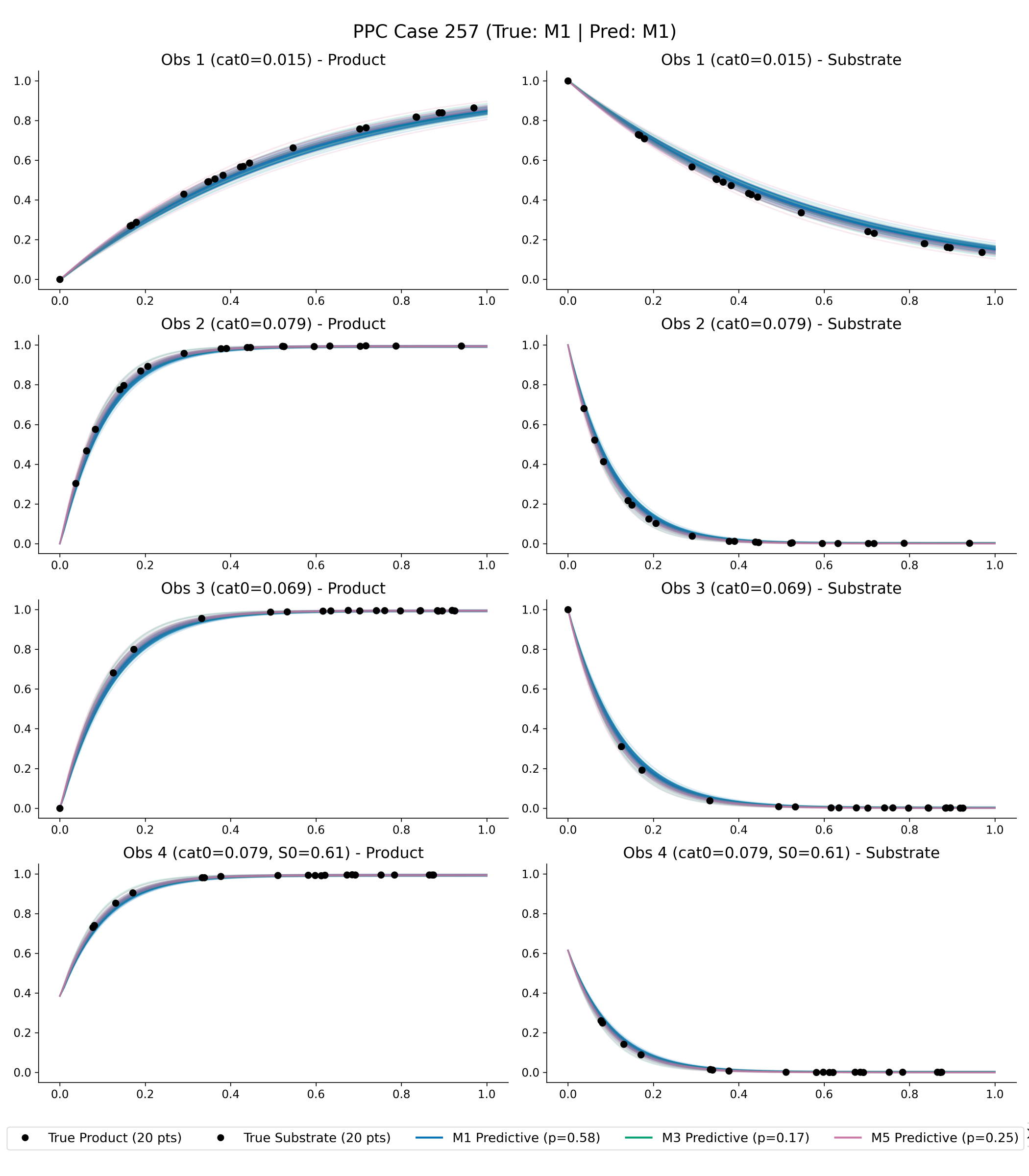}
    \caption{PPC for a test trajectory generated by mechanism $M_1$ under the high-resolution setup (1) with four candidate models. Black dots denote the $20$ discrete observations for both species across four distinct initial conditions. The solid lines represent $100$ posterior predictive trajectories sampled from the network's inferred joint posterior.}
    \label{fig:ppc_4M}
\end{figure}

\begin{figure}[h!]
    \centering
    \includegraphics[width=\linewidth]{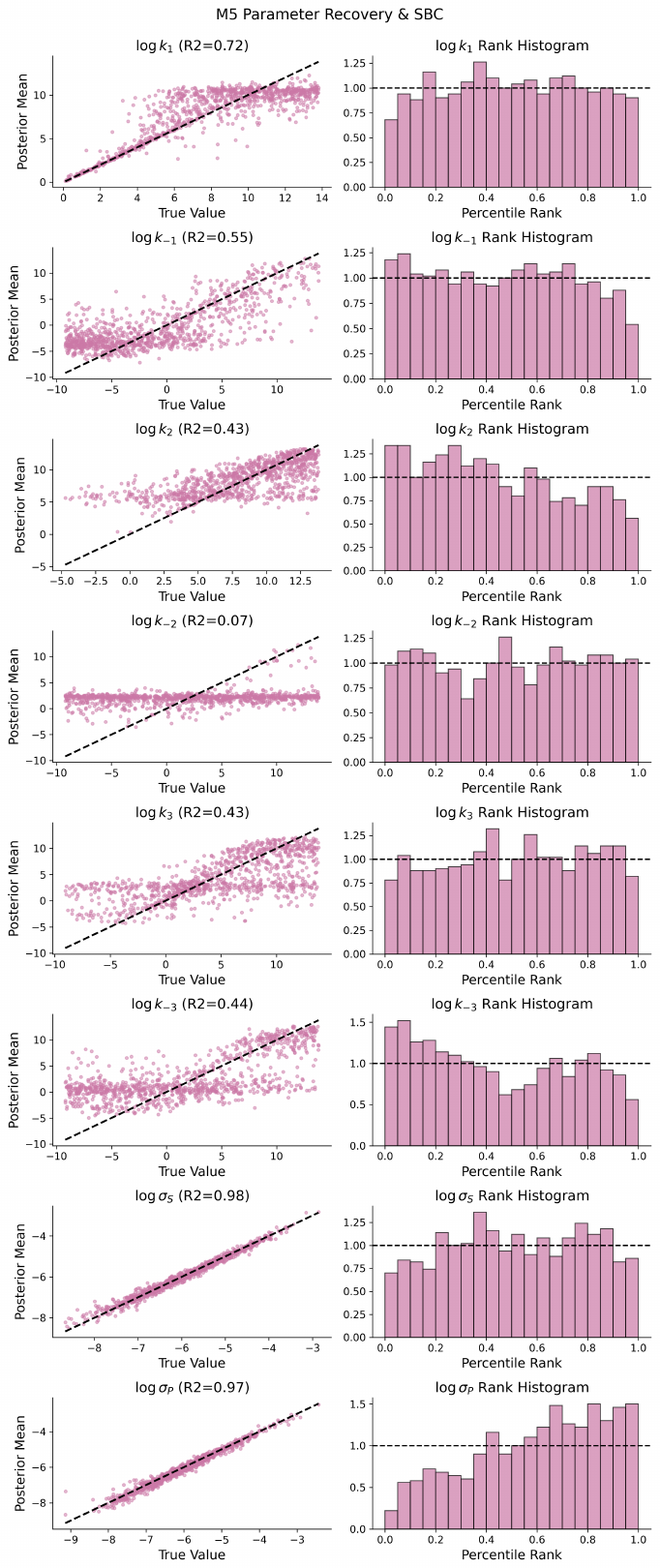}
    \caption{Inference calibration across all testing cases from $M_5$. Left Panel: Posterior mean estimates vs. the true values. Right Panel: Rank histograms for all parameters (normalized to $(0,1)$). The dashed black line indicated the level for uniform density ($1$).}
    \label{fig:rank_M5_1}
\end{figure}

This symmetric error seems to quantify the structural similarities between $M_1$ and $M_5$. Under specific parameter regimes, the concentration trajectories generated by these two mechanisms are hard to distinguish, even when both the substrate and product are observed. For example, in Figure \ref{fig:ppc_4M} we show the PPC from a test case in which both $M_1$, $M_5$, (and even $M_3$) can produce trajectories that tightly track the data. The network correctly captures the uncertainty associated with this kind of case. Moreover, all other assessments indicate inference is well-calibrated for this expanded model too. Table \ref{tab:coverage_4model} shows coverage results, Figure \ref{fig:RD_4M} the reliability diagram for the model selection posterior probabilities, and Figure \ref{fig:rank_M5_1} the rank histograms for every parameter of $M_5$ (results for all other mechanisms remained consistent with those from the three-model setting). We believe accuracy could be improved by extending the dimensions of the architecture when considering larger model spaces, but did not explore it for this work. 

\clearpage

\section{Protodeboronation Example: Details and Results}
\label{app:real_world_params}
Catalytic protodeboronation is an undesirable degradation process that frequently compromises Suzuki-Miyaura cross-coupling reactions \citep{ser_bulky_2025}. To optimize the reaction conditions and mitigate this degradation, chemists must first identify the primary kinetic pathway driving the catalyst deactivation. Using mechanistic candidates formalized by \citet{bures_organic_2023}, we formulate this as a model selection problem over three competing deactivation hypotheses: a unimolecular deactivation of the active catalyst-substrate complex ($M_{14}$), an inhibitor-induced deactivation where an external agent neutralizes the complex ($M_{15}$), and a substrate-induced deactivation where a second substrate molecule triggers the degradation ($M_{16}$). We detail this mechanisms below. 

\subsection{Mechanism 14}
Mechanism 14 is defined with any set of reaction rate constants such that at least one reaction with the reaction run $[S]_0 = 1$ and $[cat]_0$ between 3 and 7 mol\% of catalyst has between 50 and 90\% of active catalyst ($cat$ + $catS$) at 50\% of the maximum conversion.

\begin{center}
\begin{minipage}[c]{0.46\textwidth}
\centering
\begin{tikzpicture}[>=stealth, thick, scale=0.8, transform shape]
    \node (cat) at (0, 2) {cat};
    \node (catS) at (0, -2) {catS};
    \node (S) at (3, 2) {S};
    \node (P) at (-3, 2) {P};

    \coordinate (R1f) at (2.2, 0);
    \draw[-] (cat) to[bend left=15] (R1f);
    \draw[-] (S) to[bend left=15] (R1f);
    \draw[->] (R1f) to[bend left=15] node[right=0.1cm] {$k_1$} (catS);

    \coordinate (R1r) at (1.2, 0);
    \draw[dashed] (catS) to[bend right=15] (R1r) node[left=0.01cm] {$k_{-1}$};
    \draw[->, dashed] (R1r) to[bend right=15] (S);
    \draw[->, dashed] (R1r) to[bend right=15] (cat);

    \coordinate (R2f) at (-2.2, 0);
    \draw[-] (catS) to[bend left=15] node[left=0.1cm] {$k_2$} (R2f);
    \draw[->] (R2f) to[bend left=15] (cat);
    \draw[->] (R2f) to[bend right=15] (P);

    \coordinate (R2r) at (-1.2, 0);
    \draw[dashed] (cat) to[bend right=15] (R2r) node[right=0.1cm] {$k_{-2}$};
    \draw[dashed] (P) to[bend left=35] (R2r);
    \draw[->, dashed] (R2r) to[bend right=15] (catS);

    \node (inactive) at (0, -5) {inactive catS};
    \draw[->] (catS) -- node[right=0.1cm] {$k_{-3}$} (inactive);
\end{tikzpicture}
\end{minipage}
\hfill
\begin{minipage}[c]{0.53\textwidth}
\begin{equation*}
\begin{aligned}
 \frac{d[S]}{dt} &= k_{-1}[catS] - k_1[S][cat] \\
 \frac{d[P]}{dt} &= k_2[catS] - k_{-2}[cat][P] \\     
 \frac{d[cat]}{dt} &= (k_{-1} + k_2)[catS] - (k_1[S] + k_{-2}[P])[cat] \\     
 \frac{d[catS]}{dt} &= (k_1[S] + k_{-2}[P])[cat] - (k_{-1} + k_2 + k_{-3})[catS] \\     
 \frac{d[\text{inactive } catS]}{dt} &= k_{-3}[catS]
\end{aligned}
\end{equation*}
\end{minipage}
\end{center}

\subsection{Mechanism 15}
Mechanism 15 is defined with any set of reaction rate constants such that at least one reaction with the reaction run $[S]_0 = 1$ and $[cat]_0$ between 3 and 7 mol\% of catalyst has more than 10\% of inactive catalyst ($inactive catSI$) at 50\% of the maximum conversion.

\begin{center}
\begin{minipage}[c]{0.45\textwidth}
\centering
\begin{tikzpicture}[>=stealth, thick, scale=0.8, transform shape]
    \node (cat) at (0, 2) {cat};
    \node (catS) at (0, -2) {catS};
    \node (S) at (3, 2) {S};
    \node (P) at (-3, 2) {P};

    \coordinate (R1f) at (2.2, 0);
    \draw[-] (cat) to[bend left=15] (R1f);
    \draw[-] (S) to[bend left=15] (R1f);
    \draw[->] (R1f) to[bend left=15] node[right=0.1cm] {$k_1$} (catS);

    \coordinate (R1r) at (1.2, 0);
    \draw[dashed] (catS) to[bend right=15] (R1r) node[left=0.01cm] {$k_{-1}$};
    \draw[->, dashed] (R1r) to[bend right=15] (S);
    \draw[->, dashed] (R1r) to[bend right=15] (cat);

    \coordinate (R2f) at (-2.2, 0);
    \draw[-] (catS) to[bend left=15] node[left=0.1cm] {$k_2$} (R2f);
    \draw[->] (R2f) to[bend left=15] (cat);
    \draw[->] (R2f) to[bend right=15] (P);

    \coordinate (R2r) at (-1.2, 0);
    \draw[dashed] (cat) to[bend right=15] (R2r) node[right=0.1cm] {$k_{-2}$};
    \draw[dashed] (P) to[bend left=35] (R2r);
    \draw[->, dashed] (R2r) to[bend right=15] (catS);

    \node (inactive) at (0, -5) {inactive catSI};
    \node (inh) at (-2.5, -3.5) {inhibitor};
    \coordinate (deact) at (0, -3.5);
    \draw[-] (catS) -- (deact);
    \draw[-] (inh) to[bend right=15] (deact);
    \draw[->] (deact) -- node[right=0.1cm] {$k_{-3}$} (inactive);
\end{tikzpicture}
\end{minipage}
\hfill
\begin{minipage}[c]{0.54\textwidth}
\begin{equation*}
\begin{aligned}
 \frac{d[S]}{dt} &= k_{-1}[catS] - k_1[S][cat] \\
 \frac{d[P]}{dt} &= k_2[catS] - k_{-2}[cat][P] \\     
 \frac{d[cat]}{dt} &= (k_{-1} + k_2)[catS] - (k_1[S] + k_{-2}[P])[cat] \\     
 \frac{d[catS]}{dt} &= (k_1[S] + k_{-2}[P])[cat] \\
 & \quad - (k_{-1} + k_2 + k_{-3}[\text{inhibitor}])[catS] \\     
 \frac{d[\text{inhibitor}]}{dt} &= -k_{-3}[\text{inhibitor}][catS] \\
 \frac{d[\text{inactive } catSI]}{dt} &= k_{-3}[\text{inhibitor}][catS]
\end{aligned}
\end{equation*}
\end{minipage}
\end{center}

\subsection{Mechanism 16}
Mechanism 16 is defined with any set of reaction rate constants such that at least one reaction with the reaction run $[S]_0 = 1$ and $[cat]_0$ between 3 and 7 mol\% of catalyst has between 50 and 90\% of active catalyst ($cat + catS$) at 50\% of the maximum conversion.

\begin{center}
\begin{minipage}[c]{0.45\textwidth}
\centering
\begin{tikzpicture}[>=stealth, thick, scale=0.8, transform shape]
    \node (cat) at (0, 2) {cat};
    \node (catS) at (0, -2) {catS};
    \node (S) at (3, 2) {S};
    \node (P) at (-3, 2) {P};

    \coordinate (R1f) at (2.2, 0);
    \draw[-] (cat) to[bend left=15] (R1f);
    \draw[-] (S) to[bend left=15] (R1f);
    \draw[->] (R1f) to[bend left=15] node[right=0.1cm] {$k_1$} (catS);

    \coordinate (R1r) at (1.2, 0);
    \draw[dashed] (catS) to[bend right=15] (R1r) node[left=0.01cm] {$k_{-1}$};
    \draw[->, dashed] (R1r) to[bend right=15] (S);
    \draw[->, dashed] (R1r) to[bend right=15] (cat);

    \coordinate (R2f) at (-2.2, 0);
    \draw[-] (catS) to[bend left=15] node[left=0.1cm] {$k_2$} (R2f);
    \draw[->] (R2f) to[bend left=15] (cat);
    \draw[->] (R2f) to[bend right=15] (P);

    \coordinate (R2r) at (-1.2, 0);
    \draw[dashed] (cat) to[bend right=15] (R2r) node[right=0.1cm] {$k_{-2}$};
    \draw[dashed] (P) to[bend left=35] (R2r);
    \draw[->, dashed] (R2r) to[bend right=15] (catS);

    \node (inactive) at (0, -5) {inactive catS$_2$};
    \node (S2) at (-2.5, -3.5) {S};
    \coordinate (deact) at (0, -3.5);
    \draw[-] (catS) -- (deact);
    \draw[-] (S2) to[bend right=15] (deact);
    \draw[->] (deact) -- node[right=0.1cm] {$k_{-3}$} (inactive);
\end{tikzpicture}
\end{minipage}
\hfill
\begin{minipage}[c]{0.54\textwidth}
\begin{equation*}
\begin{aligned}
 \frac{d[S]}{dt} &= k_{-1}[catS] - k_1[S][cat] - k_{-3}[S][catS] \\
 \frac{d[P]}{dt} &= k_2[catS] - k_{-2}[cat][P] \\     
 \frac{d[cat]}{dt} &= (k_{-1} + k_2)[catS] - (k_1[S] + k_{-2}[P])[cat] \\     
 \frac{d[catS]}{dt} &= (k_1[S] + k_{-2}[P])[cat] \\
 & \quad - (k_{-1} + k_2 + k_{-3}[S])[catS] \\     
 \frac{d[\text{inactive } catS_2]}{dt} &= k_{-3}[S][catS]
\end{aligned}
\end{equation*}
\end{minipage}
\end{center}

\subsection{Inference Details and Results}
\label{sec:real_results}

Real-world kinetic data is limited by experimental constraints. In this application, the available observational data consists solely of product yield measurements, sampled at just $8$ time points. The experimental data was collected by \citet{ser_bulky_2025} (we take them from their Table S22) across $N_{\text{exp}} = 3$ independent setups featuring varying initial catalyst loadings ($5\%$, $10\%$, and $15\%$). This represents a regime of low observability: intermediate species---such as the active catalyst, and the degraded inactive species---are unobserved. Under these conditions, traditional likelihood-free inference methods struggle significantly due to structural ambiguity and high rejection rates.

To perform inference and mechanism selection, we trained a dedicated instance of our neural inference pipeline tailored to this three-model hypothesis space ($\mathcal{M} = \{M_{14}, M_{15}, M_{16}\}$). The architecture was trained on synthetic trajectories ($936,000$ per model) simulated from the corresponding ODEs, with only the product observed. The permutation-invariant SAB within our summary network naturally accommodates the three independent initial conditions, processing the $5\%$, $10\%$, and $15\%$ catalyst loading trajectories as an unordered set to extract global kinetic features. 

For inference, the time-normalized experimental time-series were passed through the trained architecture to compute posterior model probabilities and estimate the joint posterior distributions for the kinetic rate constants.

\begin{figure}[h!]
    \centering
    \includegraphics[width=\linewidth]{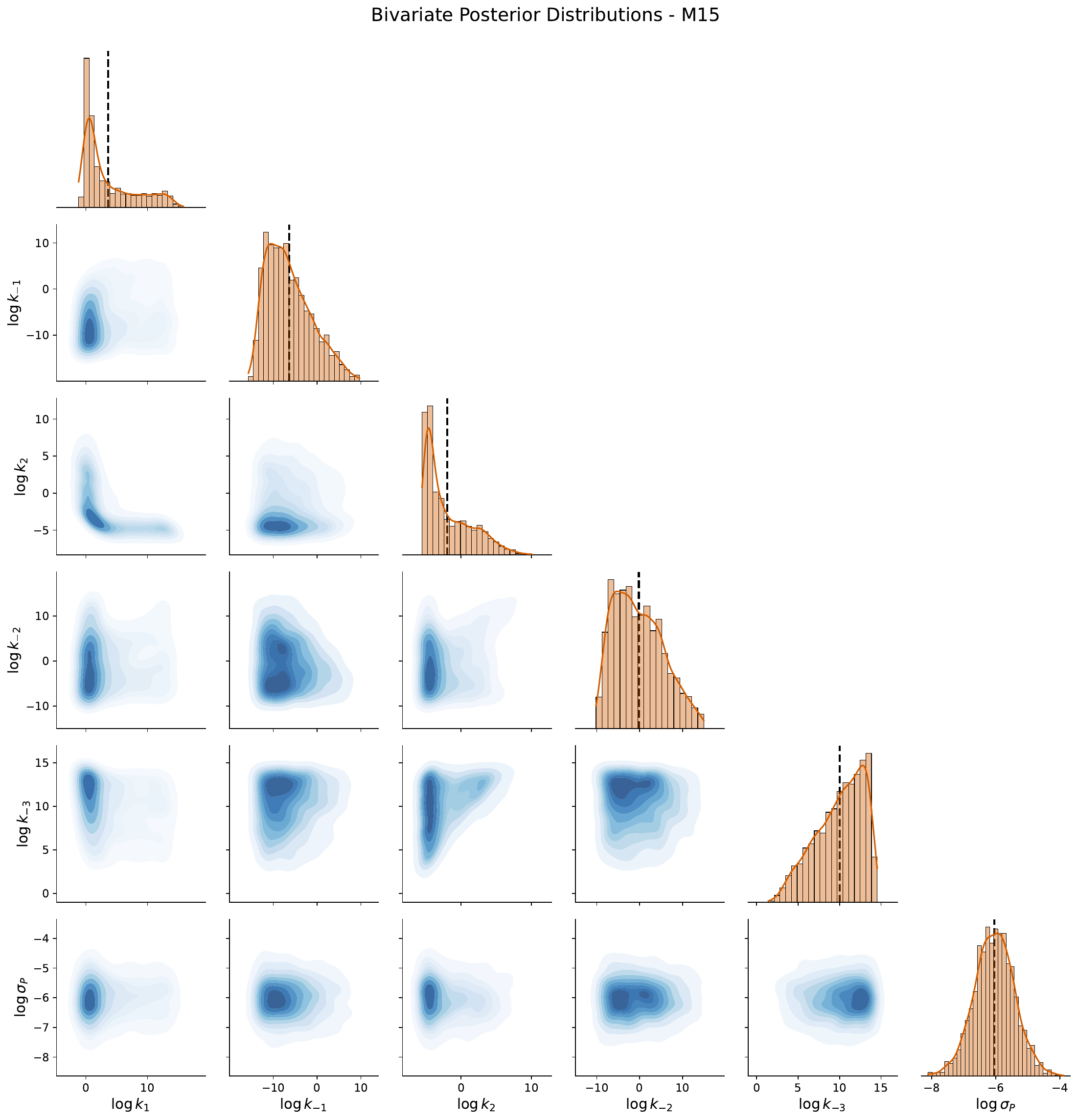}
    \caption{Bivariate and marginal posterior distributions ($2,000$ samples) for the most probable protodeboronation mechanism. The diagonal panels display the marginal posterior distributions. The lower-triangular panels feature bivariate KDE contour maps, illustrating the pairwise uncertainty between the kinetic constants.}
    \label{fig:bivariate_pos}
\end{figure}

Table \ref{tab:real_data_params} provides the marginal posterior point estimates (mean and median) for the kinetic rate constants and measurement noise across all three candidate models ($M_{14}, M_{15}, M_{16}$) evaluated in Section \ref{sec:real_world}. The estimates are provided in logarithmic physical units. To quantify the uncertainty surrounding these estimates, Figure \ref{fig:posteriors_1} visualizes the marginal posterior distributions for the highly favored inhibitor-induced pathway ($M_{15}$). Lastly, Figure \ref{fig:bivariate_pos} shows the corresponding pairwise posterior distributions. The plots show that even though the posterior distributions for some of the parameters are very wide, the posterior samples account for relationships between them.

\begin{table}[htpb]
\centering
\caption{Parameter Estimates. Posterior parameter estimates for the protodeboronation structural hypotheses.}
\label{tab:real_data_params}
\begin{tabular}{@{}lrr|rr|rr@{}}
\toprule
 & \multicolumn{2}{c}{\textbf{$M_{14}$ (3.71\%)}} & \multicolumn{2}{c}{\textbf{$M_{15}$ (92.19\%)}} & \multicolumn{2}{c}{\textbf{$M_{16}$ (4.10\%)}} \\ \cmidrule(l){2-7} 
\textbf{Parameter} & \textbf{Mean} & \textbf{Median} & \textbf{Mean} & \textbf{Median} & \textbf{Mean} & \textbf{Median} \\ \midrule
$\log k_{1}$ & $3.9572$ & $2.2728$ & $3.5822$ & $1.0382$ & $4.2962$ & $2.6632$ \\
$\log k_{-1}$ & $-5.6564$ & $-6.7500$ & $-6.1921$ & $-7.2830$ & $-5.8420$ & $-6.8342$ \\
$\log k_{2}$ & $-1.8568$ & $-3.9131$ & $-1.8872$ & $-2.5498$ & $-2.0608$ & $-4.0818$ \\
$\log k_{-2}$ & $-0.9392$ & $-1.4539$ & $-0.2301$ & $-1.2468$ & $-1.0966$ & $-1.8808$ \\
$\log k_{-3}$ & $-2.1533$ & $-4.2273$ & $9.9645$ & $10.5861$ & $2.2611$ & $0.2867$ \\
$\log \sigma_{P}$ & $-5.9537$ & $-5.9104$ & $-6.0284$ & $-6.0212$ & $-5.9492$ & $-5.9434$ \\ \bottomrule
\end{tabular}
\end{table}

\begin{figure}[h!]
    \centering
    \includegraphics[width=\linewidth]{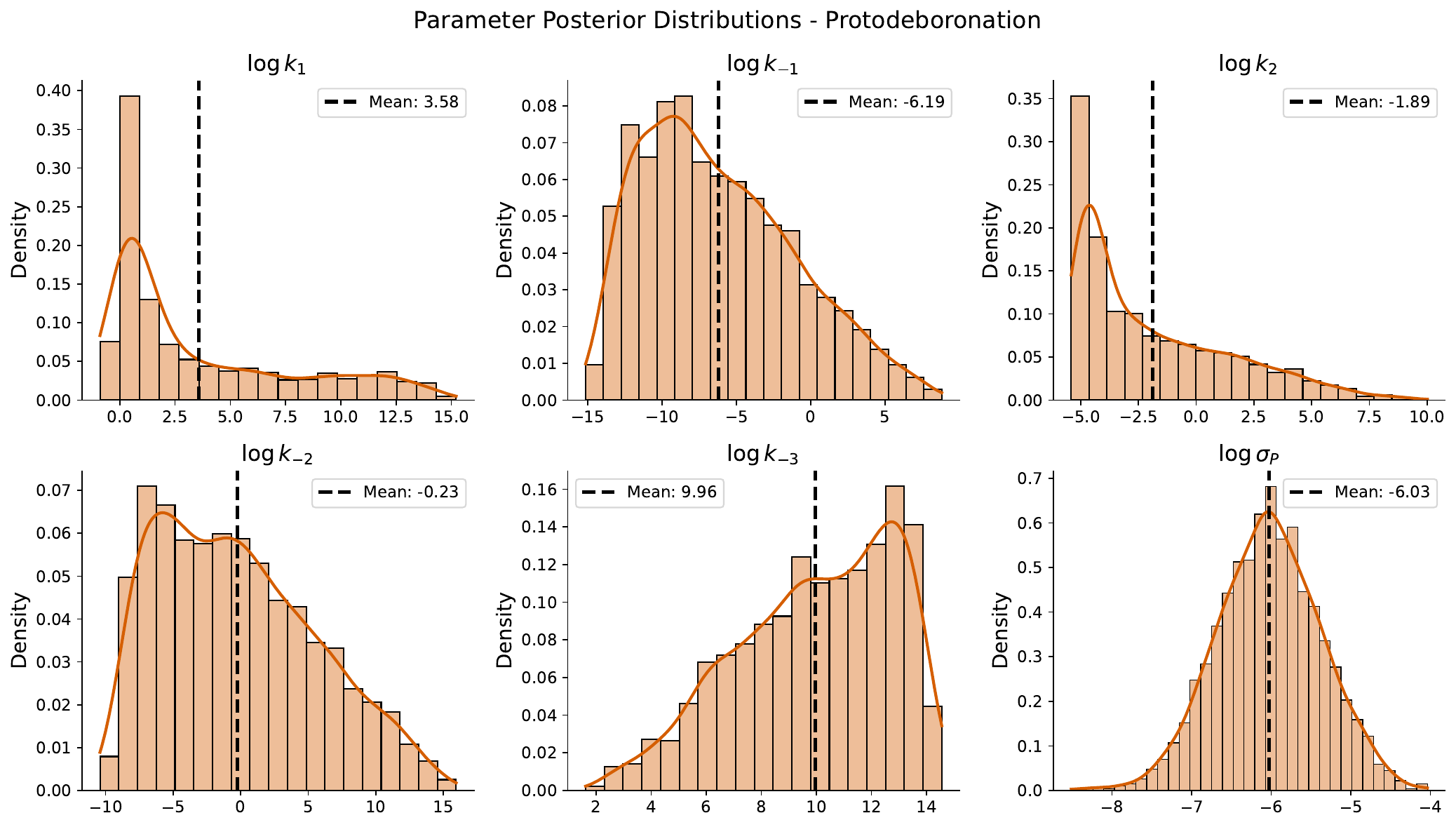}
    \caption{Estimated posterior distributions for $M_{15}$, which was estimated to be the most likely mechanism from our candidate models. We use $2,000$ samples, for all parameters estimated from real protodeboronation data. The estimated posterior mean (in the log-space) is marked with black dashed line.}
    \label{fig:posteriors_1}
\end{figure}

\end{document}